\documentclass{article}
\usepackage{iclr2027_conference,times}
\usepackage[T1]{fontenc}
\usepackage[utf8]{inputenc}
\usepackage{amsmath,amssymb,booktabs,multirow,tabularx,array,graphicx,microtype}
\usepackage{longtable}
\usepackage{tikz}
\usetikzlibrary{positioning,calc}
\usepackage{hyperref,url}
\definecolor{paperlink}{RGB}{25,63,105}

\definecolor{gapup}{RGB}{22,110,62}
\definecolor{gapdown}{RGB}{168,42,42}

\hypersetup{colorlinks=true,citecolor=paperlink,linkcolor=paperlink,urlcolor=paperlink}
\newcolumntype{Y}{>{\raggedright\arraybackslash}X}
\newcolumntype{P}[1]{>{\raggedright\arraybackslash}p{#1}}
\newcommand{\bench}{\textsc{Sanzaru}}
\newcommand{\fname}[1]{#1}
\newcommand{\abil}[1]{\emph{#1}}
\newcommand{\flag}[1]{\texttt{#1}}
\newcommand{\model}[1]{#1}

\title{Sanzaru: Benchmarking Protocol Formation\\under Epistemic \& Communication Asymmetry}
\author{Yiping Li, Zhiyu An \& Wan Du\\
Department of Computer Science and Engineering\\
University of California, Merced\\
Merced, CA 95343\\
\texttt{\{yipingli, zan7, wdu3\}@ucmerced.edu}
}
\iclrfinalcopy
\newcommand{\sanzarupreprint}{}
\newcommand{\CodedEpisodes}{771}
\newcommand{\LostEpisodes}{771}
\newcommand{\CoderAgreement}{86\%}
\newcommand{\CoderKappa}{0.83}

\newcommand{\CoderSplit}{141}
\newcommand{\CoderThreeWay}{14}
\newcommand{\CoderUnanimousN}{616}
\newcommand{\CoderJudged}{159}
\newcommand{\CoderReviewed}{157}
\newcommand{\CoderOverride}{34}
\newcommand{\FlagAssumedMeaning}{74\%}
\newcommand{\FlagUngroundedRef}{67\%}
\newcommand{\FlagOmission}{54\%}
\newcommand{\FlagWrongRound}{28\%}
\newcommand{\FlagDeadChannel}{19\%}
\newcommand{\FlagNone}{18}

\newcommand{\DeadChannelWon}{11\%}
\newcommand{\UngroundedBest}{13\%}
\newcommand{\UngroundedRest}{48\%}
\newcommand{\AssumedMeaningMin}{52\%}
\newcommand{\AssumedMeaningMax}{84\%}
\newcommand{\GeminiDeadWon}{12 of 186}

\newcommand{\ThinkOff}{20.4\%}
\newcommand{\ThinkOffPlus}{8.3\%}
\newcommand{\ThinkOffTimeout}{75}
\newcommand{\ThinkOffWrong}{97}
\newcommand{\ThinkOn}{30.6\%}

\newcommand{\ThinkOnTimeout}{41}
\newcommand{\ThinkOnWrong}{109}
\newcommand{\ThinkMax}{59.7\%}
\newcommand{\ThinkMaxPlus}{50.0\%}
\newcommand{\ThinkMaxTimeout}{60}
\newcommand{\ThinkMaxWrong}{27}

\newcommand{\TrioAll}{49\%}
\newcommand{\SoloAll}{90\%}
\newcommand{\SoloTrio}{41\%}
\newcommand{\NumModels}{7}

\newcommand{\BestAll}{86\%}

\newcommand{\WorstAll}{20\%}

\newcommand{\WeakEpisodes}{1512}
\newcommand{\WeakFailures}{771}

\newcommand{\SlowEpisodes}{18}
\newcommand{\SlowLongest}{28}
\newcommand{\SlowMedian}{5}
\newcommand{\MorseLongest}{61}

\begin{document}
\maketitle
\lhead{Preprint}

\begin{abstract}
A language model's ability to solve a task individually does not ensure that a team can solve it under unequal access to information and communication.
\bench{} examines this gap through cooperative bomb-defusal puzzles adapted from \emph{BOMBANANA!} \citep{lefto2026bombanana}.
Task information and action authority are divided among three agents: one cannot hear, another cannot speak, and the third cannot see.
The benchmark contains 120 puzzle instances across seven families, each targeting a distinct coordination ability.
Across \WeakEpisodes{} episodes, \NumModels{} models achieve an overall success rate of \TrioAll{}, ranging from \WorstAll{} to \BestAll{} under the evaluated configurations.
Teams develop or adopt diverse communication protocols, from simple gesture codes to Morse and encodings of entire manuals.
Yet trace analysis attributes about 60\% of \WeakFailures{} team losses to communication, and agents attempt to use channels unavailable to their teammates despite explicit role instructions.
Together, task outcomes and interaction traces enable evaluation of both collaborative performance and perspective taking throughout cooperation.
\ifdefined\sanzarupreprint
The benchmark is available at \url{https://github.com/markli404/sanzaru-benchmark}.
\fi
\end{abstract}

\section{Introduction}
Multi-agent language-model systems increasingly assign agents complementary roles \citep{ossowski2025comma,sun2025overcooked}.
This motivates a question: if a task can be solved by a single agent, can a team still solve it reliably?
Existing benchmarks divide information and action authority between agents while leaving communication symmetric \citep{ossowski2025comma,parekh2026gptnt}.
We ask what happens when agents must account for how their teammates can communicate.

\begin{figure}[t]
\centering
\includegraphics[width=\linewidth]{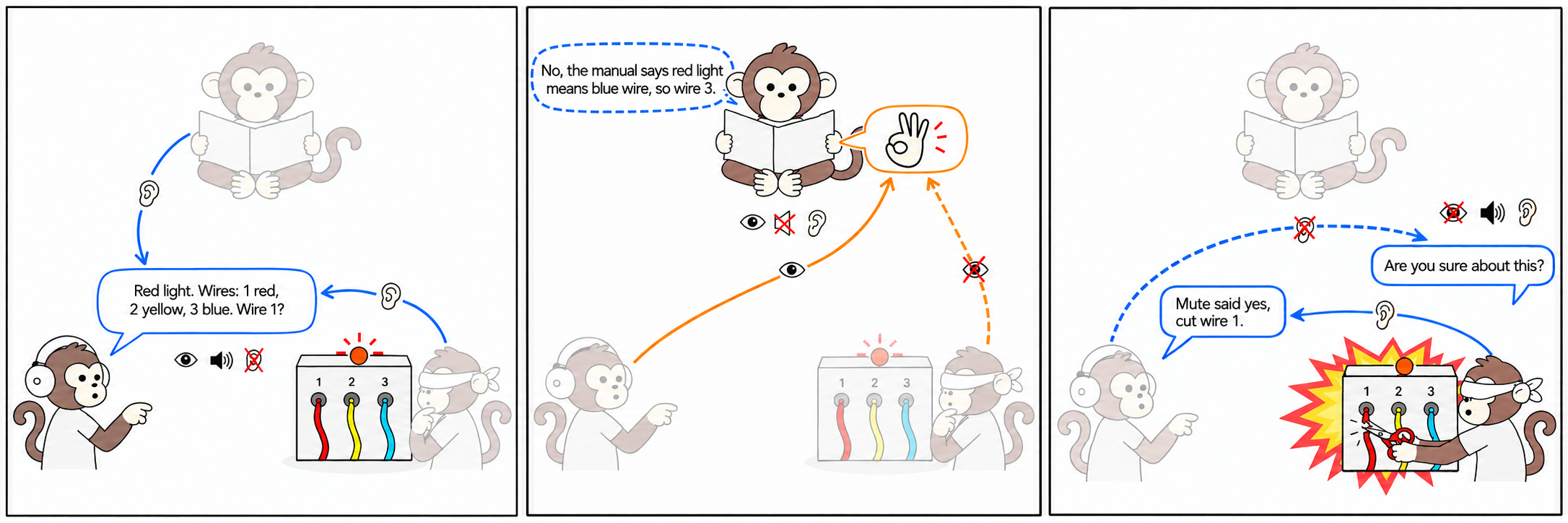}
\caption{\textbf{Three monkeys, three communication limits, one unfortunate wire.} Information is split across the team: Deaf reports what it sees, Mute consults the manual but can only gesture, and Blind makes the cut. Mute means ``wire 3,'' Deaf reads the gesture as ``yes,'' and the bomb goes off.}
\label{fig:overview}
\end{figure}

We introduce \bench{}, adapted from the cooperative bomb-defusal game \emph{BOMBANANA!} \citep{lefto2026bombanana}.
\bench{} is a series of three-role bomb-defusal puzzles in which each agent has a unique set of observation and communication restrictions.
The goal is to defuse a bomb by coordinating information and actions across three agents.
The Deaf agent can see the bomb but cannot hear; the Mute agent holds the defusal manual but can only communicate through gestures; and the Blind agent can manipulate the bomb but cannot see (Figure~\ref{fig:overview}).

The challenge is not merely to combine their knowledge, but to establish a communication protocol that allows the right information to reach the right agent in a form that agent can receive and act upon.

This setting connects collaboration to two strands of research.
The first concerns persona consistency and following role prompts.
Work in this area evaluates whether elicited responses express the intended persona.
PersonaLLM combines personality questionnaires with writing tasks, while PersonaGym assesses responses to persona-specific situational questions \citep{jiang2024personallm,samuel2025personagym}.
\citet{gonnermann2026maintaining} find that stable self-reported personas can coexist with declining observer-rated expression during extended conversations.
A plausible answer about one's role does not establish that the role will govern subsequent decisions.
We call this \emph{functional persona consistency}, whether an agent's decisions respect its assigned information, capabilities, and responsibilities throughout a task.
\bench{} examines this property through attempted messages and actions in a simulator that enforces the role restrictions.

The second strand concerns theory of mind (ToM) in language models, reasoning about others' knowledge and beliefs.
FANToM probes information-asymmetric conversations through questions about their participants \citep{kim2023fantom}.
SimpleToM distinguishes mental-state inference from behavior prediction and judgment, showing that success at the former need not transfer to the latter \citep{gu2026simpletom}.
Answering questions about others still differs from making one's own actions depend on their perspectives.
Interactive work such as SOTOPIA-ToM studies information management during multi-agent exchanges \citep{ys2026sotopiatom}.
In \bench{}, what a teammate receives depends on both their role and the communication channel.
We examine whether agents adapt their communication and actions to what their teammates know and can receive.
This lets us study whether perspective-taking guides cooperation as teams develop signaling conventions throughout the task.

\bench{} comprises seven puzzle families spanning 24 variants and 120 seeded instances.
We evaluate \NumModels{} models from major providers, with success rates ranging from \WorstAll{} to \BestAll{}.
With all views, the manual, and action authority, \model{GPT-5.6-Luna} solves \SoloAll{} of the puzzles alone, compared with \SoloTrio{} for its three-agent team on the same puzzles.
We analyze all \LostEpisodes{} failed episodes to identify where cooperation breaks down and how agents handle their teammates' limited access to information and communication.
Failure-mode coding attributes about 60\% of the lost episodes to communication.

We make three contributions:
\begin{enumerate}
\setlength{\itemsep}{3pt}\setlength{\parskip}{0pt}\setlength{\leftmargin}{0pt}
\item \textbf{A benchmark for protocol formation under epistemic and communication asymmetry.}
Seven diagnostic families combine private information, complementary capabilities, and restrictions on message delivery in stateful puzzles.
\item \textbf{An operational evaluation of functional persona consistency and perspective taking.}
We examine whether agents' messages and actions account for their own roles and their teammates' information and capabilities throughout cooperation.
\item \textbf{An empirical analysis of collaboration failures.}
A seven-model evaluation, a solo comparison, eight failure modes, and five perspective-taking flags characterize where teams struggle to coordinate.
\end{enumerate}

\section{Related work}
\label{sec:related}

\textbf{Collaboration under information asymmetry.} Several closely related benchmarks require agents to communicate across private views.
COMMA pairs a multimodal solver with a manual-holding expert to solve collaborative puzzles \citep{ossowski2025comma}.
GPTNT places a defuser and an expert in the original \emph{Keep Talking and Nobody Explodes} game, where they communicate asynchronously against a live countdown \citep{parekh2026gptnt}.
Collab-Overcooked divides resources and task knowledge between agents who coordinate through language \citep{sun2025overcooked}.
Two recent benchmarks more directly study collaboration under asymmetric information.
\citet{peng2025verification} place two agents with different knowledge and skills in a tabletop puzzle that requires reasoning, communication, and action, while \citet{wang2026deliberative} study joint decision making under partial and asymmetric observations.
Game-based work also examines visual and conversational grounding through dialogue \citep{hakimov2025grounding}.
Unlike these benchmarks, \bench{} makes communication asymmetric, so task information must be adapted to each recipient and relayed to the agent who can act.

\textbf{Communication conventions.} Emergent communication studies train agents to coordinate through learned messages, including single-message referential games \citep{lazaridou2017emergence} and referential games that span modalities and allow bidirectional exchange \citep{evtimova2018emergent}.
\citet{ashery2025conventions} study how conventions spread through repeated interactions in decentralized LLM populations.
\bench{} instead evaluates fixed pretrained agents within a single episode, without parameter updates or shared memory across episodes.
They may establish a signaling convention during the task, but must also use it throughout the episode.
Messages that appear informative need not affect a partner's behavior \citep{lowe2019pitfalls}.
We therefore examine attempted messages, what each recipient receives, and the actions that follow, alongside task success.

\textbf{Perspective taking in coordinated action.}
Prior work tends to evaluate perspective taking and coordinated action separately.
FANToM tests whether a model can infer what participants in an information-asymmetric conversation know or believe, but scores these inferences through explicit answers rather than through the decisions that follow from them \citep{kim2023fantom}.
Multi-agent benchmarks instead evaluate coordination itself, often under partial observability \citep{bard2020hanabi,agashe2025coordination,ruan2025swarmbench,park2026pacbench}.
Some restrict communication, but apply the same restrictions to every agent rather than tying them to a role.
\bench{} combines these two demands by requiring agents to track their teammates' knowledge, perceptual access, and communication capabilities.
Because information must be reformulated for each recipient and relayed to the actor, the benchmark measures perspective taking through its effect on communication and action rather than through a model's report of another agent's mental state.

\section{Benchmark design}
\label{sec:method}
\subsection{The defusal game}
An episode of \bench{}, adapted from the party game \emph{BOMBANANA!} \citep{lefto2026bombanana}, asks three agents to defuse a bomb.
The bomb displays observable features, and a manual maps those features to the one safe wire.
A wrong cut fails the episode.
The information and authority needed to identify and cut that wire are split across the team.
Mute holds the manual but cannot speak; Blind holds the cutters but cannot see; Deaf can see and speak but cannot hear (Figure~\ref{fig:overview}).
No agent or pair can finish alone, so the solution must pass through all three roles, each with its own view, its own channels and its own permitted actions (Table~\ref{tab:roles}).

\begin{table}[t]
\centering\scriptsize
\caption{Role contracts in the standard team, shown on a \fname{Base} instance. Observations, deliveries, and calls are verbatim text; gesture symbols are emoji glyphs in the implementation, named here in print.}
\label{tab:roles}
\begin{tabularx}{\linewidth}{@{}P{0.09\linewidth}P{0.28\linewidth}P{0.24\linewidth}Y@{}}
\toprule
 & Deaf & Mute & Blind \\
\midrule
Perceives &
\texttt{DISPLAY [0]}\newline\texttt{LIGHT [green]}\newline\texttt{WIRES 1:[yellow]}\newline\texttt{\phantom{WIRES }2:[green]}\newline\texttt{\phantom{WIRES }3:[red]} &
{\ttfamily\begin{tabular}[t]{@{}l|l@{}}light & wire\\\hline green & red\\ red & yellow\\ yellow & green\end{tabular}} &
\texttt{DISPLAY [unreadable]}\newline\texttt{LIGHT [unreadable]}\newline\texttt{WIRES 1:[unreadable] 2:[unreadable] 3:[unreadable]} \\
\midrule
Acts &
\texttt{wait()} &
\texttt{wait()} &
\texttt{act(op="cut", position=3)}\newline\texttt{wait()} \\
\midrule
Sends &
\texttt{speak(text="Light}\newline\texttt{\phantom{sp}is green.")}\newline\texttt{gesture(to="mute",}\newline\texttt{\phantom{ge}symbols=[thumbs\_up])} &
\texttt{gesture(to="deaf",}\newline\texttt{\phantom{ge}symbols=[point\_up,}\newline\texttt{\phantom{gesymbols=}fist])} &
\texttt{speak(to="mute", text=}\newline\texttt{\phantom{sp}"I feel three wires.")} \\
\midrule
Receives &
\texttt{Mute: [victory,}\newline\texttt{\phantom{Mute: }thumbs\_up]} &
\texttt{Deaf: "Light is green."}\newline\texttt{Blind: "I feel}\newline\texttt{\phantom{Blind: }three wires."} &
\texttt{Deaf: "Cut wire 3."}\newline\texttt{Mute: [? ?]} \\
\bottomrule
\end{tabularx}
\vspace{-1em}
\end{table}

The bomb draws from a small set of parts, including a display, a light, wires, an engraved plate, a keypad, and printed props.
Only the display, light, and wires appear on every bomb.
The manual varies by bomb and may include a lookup table, describe a mechanism such as a cycling light, or warn against a printed offer.

Agents interact through four calls, \texttt{speak}, \texttt{gesture}, \texttt{act}, and \texttt{wait}; each role permits a different subset.
A gesture contains one or more symbols from an alphabet of 42 hand emoji.
Speech carries free text, including numerals and punctuation.
Gesture symbols may also be embedded in speech, allowing its spoken and visual content to reach different recipients.
Both channels can be broadcast or aimed at a single teammate.

\subsection{Seven diagnostic families}
The \fname{Base} family tests whether a fact can travel from the agent that observes it to the agent that acts.
The other six vary that loop one condition at a time, so a drop in score locates where cooperation breaks down.
The bank holds 24 variants of these families, 16 Core and 8 Plus, with five seeded instances each and 120 in all.
A Core variant changes one variable along its family's axis, making changes in performance diagnostic of that axis.
Together, the Core variants show how far along the axis a team can perform.
Plus variants combine changes to separate the strongest teams.
Appendix~\ref{app:configs} gives each variant's rules and budget.

\begin{table}[t]
\centering\small
\caption{\textbf{Seven families, one ability each.} Each of the six additional families changes one condition while holding the roles and channels fixed. The second column describes the mechanic and the third names the ability it isolates. \textbf{Table~\ref{tab:demands} states what each family asks of a team.} Appendix~\ref{app:configs} lists all 24 variants.}
\label{tab:axes}
\begin{tabularx}{\linewidth}{@{}P{0.135\linewidth}YP{0.135\linewidth}@{}}
\toprule
Family & What the bomb does & Ability \\
\midrule
\fname{Base} & The color of the light names the wire, and nothing about the bomb changes during the episode. & \abil{Relay} \\
\addlinespace[3pt]
\fname{Carving} & The wire is marked with an engraved glyph that has no common name. & \abil{Grounding} \\
\addlinespace[3pt]
\fname{Flashing Light} & The light changes color every round, and a cut is judged by what it shows at that moment. & \abil{Timing} \\
\addlinespace[3pt]
\fname{Flashback} & The light flashes a sequence once, and those rounds later scroll out of the chat history. & \abil{Memory} \\
\addlinespace[3pt]
\fname{Shuffle Wires} & The surviving wires rotate after every correct cut. & \abil{Staleness} \\
\addlinespace[3pt]
\fname{Haystack} & Displays, lights, and props change for no reason, and touching any of them is fatal. & \abil{Triage} \\
\addlinespace[3pt]
\fname{Bait} & The environment prints instructions that conflict with the manual, either inviting a fatal action or forbidding the required one. & \abil{Discipline} \\
\bottomrule
\end{tabularx}
\end{table}

\subsection{The simulator}
The simulator updates the bomb, routes messages, and determines the outcome of each episode.

The bomb can change at two points in a round.
At the start, round-dependent state advances, such as a light moving to its next color.
An action may then change the bomb when it lands.
The simulator applies three deterministic termination conditions.
The episode succeeds when the last required operation lands, fails immediately after a wrong action, or fails when its round budget expires.
Budgets run from 8 to 24 rounds, and agents see which round they are in but are never told the limit.
Caution therefore carries a cost because a team can reach the limit without acting.

By default, agents act sequentially within each round.
A message to an agent later in the order arrives in the same round; a message to an agent that has already acted arrives in the next.
The simulator samples the order at the start of an episode and holds it fixed across rounds.
Two families restrict the possible orders, as detailed in Appendix~\ref{app:configs}.

\begin{figure}[t]
\centering
\definecolor{deafc}{RGB}{29,78,137}
\definecolor{mutec}{RGB}{155,93,10}
\definecolor{blindc}{RGB}{20,95,70}
\definecolor{sysg}{RGB}{125,125,132}
\begin{tikzpicture}[
  font=\scriptsize,
  bar/.style={text width=3.75cm, align=left, rounded corners=3pt,
              inner xsep=4pt, inner ysep=4pt, text=white},
  box/.style={text width=3.75cm, align=left, font=\tiny\ttfamily, rounded corners=3pt,
              inner xsep=4pt, inner ysep=4pt},
  bub/.style={text width=3.3cm, align=left, font=\tiny\ttfamily, rounded corners=4pt,
              inner xsep=4pt, inner ysep=3.5pt},
  note/.style={text width=3.75cm, align=left, font=\tiny\itshape,
               inner xsep=4pt, inner ysep=2pt, text=black!60},
  lbl/.style={font=\tiny\scshape, text=sysg},
]
\def\colw{4.75}

\node[bar, fill=deafc]  (sD) at (0,0)       {\textbf{Deaf}};
\node[bar, fill=mutec]  (sM) at (\colw,0)   {\textbf{Mute}};
\node[bar, fill=blindc] (sB) at (2*\colw,0) {\textbf{Blind}};

\node[box, draw=deafc!45, fill=deafc!5, line width=.5pt, below=1.6mm of sD.south, anchor=north] (vD)
  {DISPLAY [8]\\ LIGHT [green]\\ WIRES 1:[red] 2:[yellow]\\ \phantom{WIRES }3:[blue]};
\node[box, draw=mutec!45, fill=mutec!5, line width=.5pt, below=1.6mm of sM.south, anchor=north] (vM)
  {DEFUSAL MANUAL\\[1pt]\setlength{\tabcolsep}{1.5pt}%
   \begin{tabular}{@{}r|lll@{}}
   \tiny d$\backslash$l & blue & green & yellow\\\hline
   4 & blue & yellow & red\\
   7 & red & blue & yellow\\
   8 & yellow & red & blue\\
   \end{tabular}};
\node[box, draw=blindc!45, fill=blindc!5, line width=.5pt, below=1.6mm of sB.south, anchor=north] (vB)
  {DISPLAY [unreadable]\\ LIGHT [unreadable]\\ WIRES 1:[unreadable] 2:[unreadable] 3:[unreadable]\\ \phantom{x}};

\node[lbl, below=1.4mm of vD.south west, anchor=north west] (pD) {your turn 1, replayed};
\node[lbl, below=1.4mm of vM.south west, anchor=north west] (pM) {your turn 1, replayed};
\node[lbl, below=1.4mm of vB.south west, anchor=north west] (pB) {your turn 1, replayed};

\node[note, below=0.4mm of pD.south west, anchor=north west] (rD)
  {Display 8, green light, wires red, yellow, blue. Sending all of it to both teammates.};
\node[note, below=0.4mm of pM.south west, anchor=north west] (rM)
  {(no text, the round resolved to a wait)};
\node[note, below=0.4mm of pB.south west, anchor=north west] (rB)
  {My gesture used a symbol outside the alphabet and was rejected. Resending.};

\node[lbl, below=1.2mm of rD.south west, anchor=north west] (iD) {arrived};
\node[lbl, below=1.2mm of rM.south west, anchor=north west] (iM) {arrived};
\node[lbl, below=1.2mm of rB.south west, anchor=north west] (iB) {arrived};

\node[bub, fill=black!6, below=0.6mm of iD.south west, anchor=north west] (mD) {(empty)};
\node[bub, fill=black!6, below=0.6mm of iM.south west, anchor=north west] (mM1)
  {Blind to Mute: Tell me the manual instructions, and which wire to cut.};
\node[bub, fill=black!6, below=0.6mm of mM1.south west, anchor=north west] (mM2)
  {Deaf says: The display reads 8. The light is green. Wires 1 red, 2 yellow, 3 blue.};
\node[bub, fill=black!6, below=0.6mm of iB.south west, anchor=north west] (mB1)
  {Deaf says: The display reads 8. The light is green. Wires 1 red, 2 yellow, 3 blue.};
\node[bub, fill=black!6, below=0.6mm of mB1.south west, anchor=north west] (mB2)
  {Mute gestures to Blind: [unreadable]};

\node[lbl, below=2.6mm of mD.south west,  anchor=north west] (oD) {you write};
\node[lbl, below=2.6mm of mM2.south west, anchor=north west] (oM) {you write};
\node[lbl, below=2.6mm of mB2.south west, anchor=north west] (oB) {you write};

\node[note, below=0.4mm of oD.south west, anchor=north west] (tD)
  {No new information, so I will ask the manual holder for the exact procedure.};
\node[note, below=0.4mm of oM.south west, anchor=north west] (tM)
  {Display 8 with green maps to red, which is position 1. Blind cannot see gestures but can count the blank slots, so I encode the position as one symbol.};
\node[note, below=0.4mm of oB.south west, anchor=north west] (tB)
  {Mute's gesture is unavailable to me, so I need the manual rule in words before cutting.};

\node[bub, fill=deafc!12, draw=deafc!45, below=0.7mm of tD.south east, anchor=north east] (cD)
  {speak(to="mute",\\ \phantom{sp}text="Please send the\\ \phantom{sptext="}exact manual\\ \phantom{sptext="}instruction for\\ \phantom{sptext="}Blind.")};
\node[bub, fill=mutec!12, draw=mutec!45, below=0.7mm of tM.south east, anchor=north east] (cM)
  {gesture(to="blind",\\ \phantom{ge}symbols=[thumbs\_up])};
\node[bub, fill=blindc!12, draw=blindc!45, below=0.7mm of tB.south east, anchor=north east] (cB)
  {speak(to="mute",\\ \phantom{sp}text="I cannot see your\\ \phantom{sptext="}gesture. State the\\ \phantom{sptext="}rule in words.")};
\end{tikzpicture}
\caption{\textbf{Per-agent conversation histories.} One round of \texttt{base-2\#2} is shown. Text is abridged but otherwise verbatim; italics mark private reasoning not delivered to teammates.}
\label{fig:turn}
\vspace{-1em}
\end{figure}
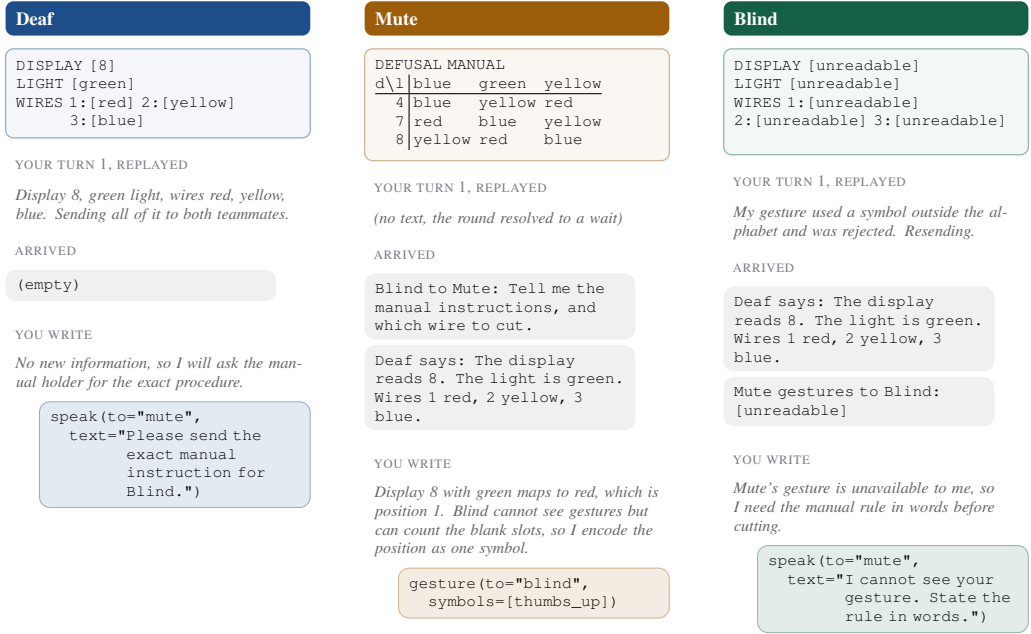

Each agent has a chat history that accumulates throughout the episode.
It begins with a system prompt describing the game and that agent's role.
At the start of every round, the simulator reports the bomb state available to that agent and any messages received since its previous turn.
The agent responds with private reasoning in plain prose and calls in a code block, and the simulator reads those calls literally rather than executing them (Figure~\ref{fig:turn}).

A call is invalid if a gesture uses a symbol outside the alphabet or speech contains a mark that cannot be pronounced.
The simulator requests a correction up to three times, after which the turn becomes a wait.
A valid call may still be undeliverable, as when Blind speaks to Deaf.
The simulator drops such a call and notifies the sender at the start of the next round.

\subsection{Hints}
We introduced three hint levels to help agents begin coordinating.
In a \model{GPT-5.6-Luna} ablation, none improved success over the no-hint condition (Appendix~\ref{app:hints}).
All three express the same principles at different levels of specificity.
Weak states them abstractly, Strong describes what the team needs, and Playbook names the roles and prescribes a procedure.
\begin{enumerate}\setlength{\itemsep}{0pt}\setlength{\parskip}{0pt}
\item A message reaches a teammate only through a sense it has, and whatever that sense cannot carry is stripped on arrival.
\item The answer reaches the hands only through whoever can read signs and speak.
\item A sign carries no meaning until the team gives it one.
\end{enumerate}
The hints provide no puzzle-specific information, and the levels differ only in how explicitly they describe the communication structure.

\section{Experiments}
\label{sec:eval}
\label{sec:results}
\textbf{Evaluation setup.} We evaluate seven models from major AI providers, including the Gemini, GLM, GPT, DeepSeek, Claude, and Qwen families.
For each model, we run the first three instances of every variant three times, using order seeds 1, 2, and 3.
This yields 216 episodes per model and hint level.
Each team is homogeneous, with all three agents using the same model but maintaining separate conversation histories.
Runs use sequential turns, temperature 0.6, a cap of 8{,}000 tokens per response, and the function-call interface.
To stay within the evaluation budget, we disabled thinking for three models; Table~\ref{tab:results} marks which models ran with thinking enabled.
Appendix~\ref{app:thinking} measures the size of this factor on \model{Qwen-3.8-Flash}, where success moves from \ThinkOff{} to \ThinkMax{} between disabled and uncapped thinking.
The overall accuracy in Table~\ref{tab:results} is therefore what each system reached as we ran it, not the model's maximum capacity.

\textbf{Success metric.} We report success as the share of episodes defused.
A family score pools all variants within that family.
The overall score weights variants equally rather than families because each family contains between two and five variants.

\textbf{Results.} Overall success ranges from 86\% for \model{Gemini-3.8-Flash} to 20\% for \model{Qwen-3.8-Flash} under Weak hints (Table~\ref{tab:results}).
\begin{table}[t]
\centering\small
\caption{\textbf{Success rate (\%) under Weak hints.} Family columns combine the Core and Plus variants of that family, as defined in Section~\ref{sec:method}. A check mark indicates that hidden thinking mode was enabled.}
\label{tab:results}
\setlength{\tabcolsep}{2.3pt}
\begin{tabular}{@{}l|c|rrrrrrr|rrr@{}}
\toprule
Model & \multicolumn{1}{c|}{Think} & \multicolumn{1}{c}{\begin{tabular}[b]{@{}c@{}}Base\end{tabular}} & \multicolumn{1}{c}{\begin{tabular}[b]{@{}c@{}}Carving\end{tabular}} & \multicolumn{1}{c}{\begin{tabular}[b]{@{}c@{}}Flashing\\Light\end{tabular}} & \multicolumn{1}{c}{\begin{tabular}[b]{@{}c@{}}Flashback\end{tabular}} & \multicolumn{1}{c}{\begin{tabular}[b]{@{}c@{}}Shuffle\\Wires\end{tabular}} & \multicolumn{1}{c}{\begin{tabular}[b]{@{}c@{}}Haystack\end{tabular}} & \multicolumn{1}{c|}{\begin{tabular}[b]{@{}c@{}}Bait\end{tabular}} & \multicolumn{1}{c}{Core} & \multicolumn{1}{c}{Plus} & \multicolumn{1}{c}{All} \\
\midrule
Gemini-3.8-Flash & \checkmark & \textbf{97.8} & \textbf{100.0} & \textbf{77.8} & 37.0 & \textbf{100.0} & \textbf{92.6} & \textbf{100.0} & \textbf{91.7} & \textbf{75.0} & \textbf{86.1} \\
GLM-5.3-Flash & \checkmark & 64.4 & 50.0 & 48.9 & \textbf{48.1} & 59.3 & 77.8 & 92.6 & 67.4 & 52.8 & 62.5 \\
GPT-5.6-Terra & \checkmark & 75.6 & 38.9 & 15.6 & 29.6 & 92.6 & 51.9 & 88.9 & 65.3 & 34.7 & 55.1 \\
DeepSeek-V4.1-Flash & -- & 71.1 & 61.1 & 20.0 & 22.2 & 81.5 & 33.3 & 63.0 & 55.6 & 36.1 & 49.1 \\
GPT-5.6-Luna & \checkmark & 60.0 & 44.4 & 11.1 & 33.3 & 81.5 & 29.6 & 33.3 & 50.0 & 22.2 & 40.7 \\
Claude-Sonnet-5 & -- & 57.8 & 5.6 & 4.4 & 22.2 & 44.4 & 22.2 & 37.0 & 34.0 & 19.4 & 29.2 \\
Qwen-3.8-Flash & -- & 26.7 & 16.7 & 6.7 & 11.1 & 40.7 & 7.4 & 37.0 & 26.4 & 8.3 & 20.4 \\
\bottomrule
\end{tabular}
\vspace{-2em}
\end{table}

\section{Analysis}
\label{sec:analysis}
\subsection{Which abilities separate the models?}

\begin{figure}[t]
\centering
\includegraphics[width=\linewidth]{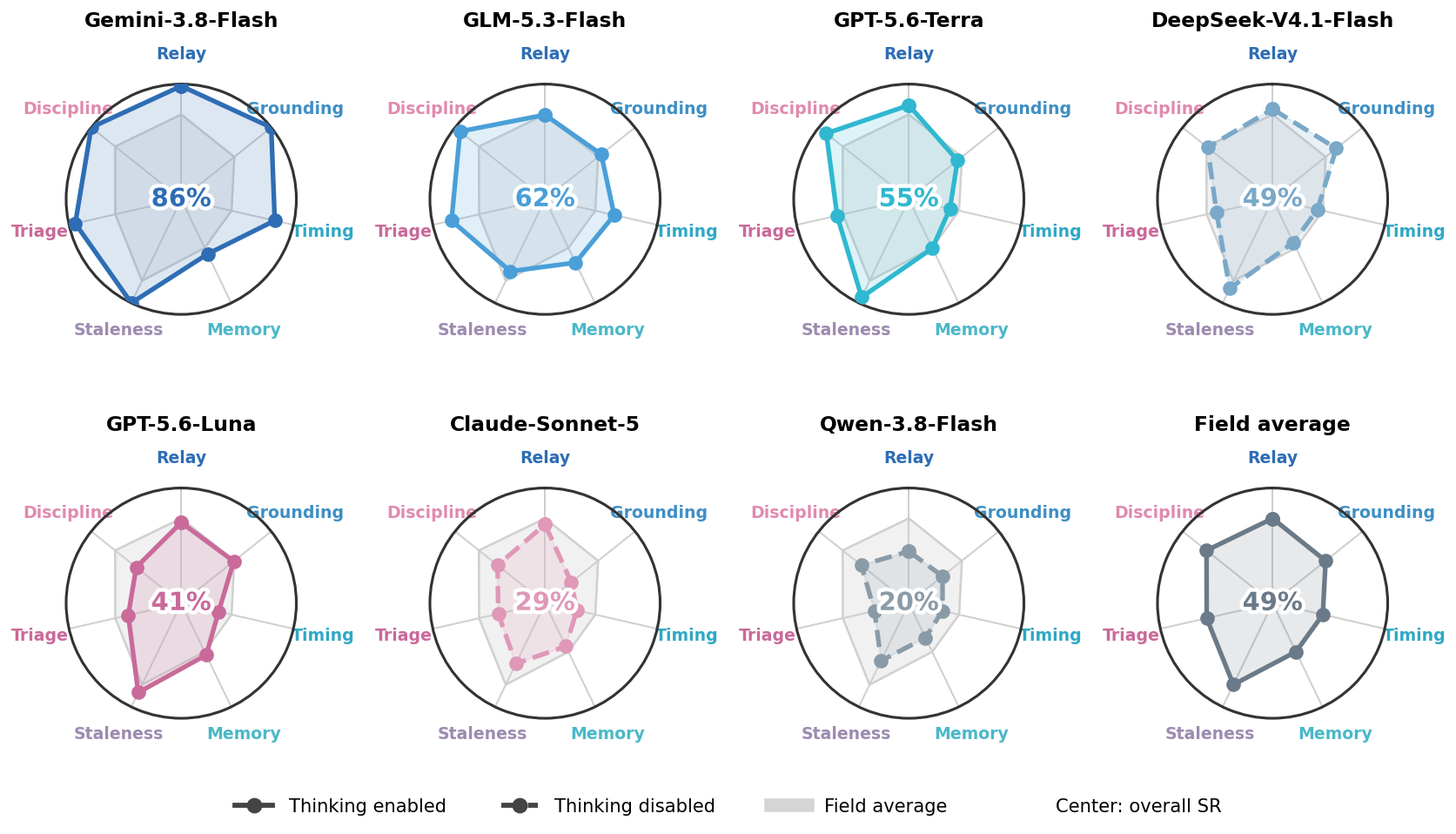}
\caption{\textbf{Capability profile under Weak hints.} One axis per family, in the order of Table~\ref{tab:axes} and labeled with the ability that table names. The grey shape behind each model is the field average, and the number at the centre is that model's overall success rate.}
\label{fig:radar}
\end{figure}

The abilities defined in Table~\ref{tab:axes} differ in both difficulty and how sharply they separate the models (Figure~\ref{fig:radar}).
\abil{Grounding} has the widest range, from 6\% to 100\%, followed closely by \abil{Triage}.
\abil{Staleness} has the highest average success and \abil{Timing} the lowest, while \abil{Memory} is the only ability on which every model remains below 50\%.
Both \abil{Staleness} and \abil{Memory} require teams to handle state changes.
They differ in whether the information needed after the change remains visible.
For every model, either \abil{Memory} or \abil{Timing} is the weakest ability; performance on the other abilities varies by model.

\subsection{How much of the difficulty is coordination?}
We test the effect of asymmetries on \model{GPT-5.6-Luna} by giving one agent every view, the manual, and the cutters.
On the same bombs, Solo solves \SoloAll{}, compared with \SoloTrio{} for the Trio (Table~\ref{tab:solo}).
The Trio loses primarily while coordinating across these asymmetries, especially on harder variants.

\fname{Flashback} is the only Core family where Solo gains nothing.
The manual names the wire by how many times the light flashed one color, but the history window drops the early rounds before that count is needed.
The agent is expected to restate the running count each round so that it stays inside the window.
Every Solo loss is a count the agent never had, either because it cut while the light was still flashing or because it counted from a history that had already lost the start.
For Luna, the remaining limit is therefore \abil{Memory} rather than coordination, and a model that kept the count would clear the family alone.

\begin{table}[t]
\centering\small
\caption{\textbf{Coordination cost by family.} Both settings use \model{GPT-5.6-Luna}. Trio is the standard team of three agents playing under Weak hints; Solo is a single agent that holds every view, the manual and the cutters, and plays without hints. Entries are the percentage of episodes defused.}
\label{tab:solo}
\setlength{\tabcolsep}{3pt}
\begin{tabular*}{\linewidth}{@{\extracolsep{\fill}}llrrrrrrrr@{}}
\toprule
 & & Base & Carving & \begin{tabular}[b]{@{}c@{}}Flashing\\Light\end{tabular} & Flashback & \begin{tabular}[b]{@{}c@{}}Shuffle\\Wires\end{tabular} & Haystack & Bait & All \\
\midrule
Core & Trio & 74.1\% & 44.4\% & 27.8\% & 44.4\% & 83.3\% & 38.9\% & 33.3\% & 50.0\% \\
 & Solo & 96.3\% & 100.0\% & 100.0\% & 44.4\% & 100.0\% & 100.0\% & 100.0\% & 92.4\% \\
 & \textbf{Gap} & \textbf{+22.2} & \textbf{+55.6} & \textbf{+72.2} & \textbf{+0.0} & \textbf{+16.7} & \textbf{+61.1} & \textbf{+66.7} & \textbf{+42.4} \\
\midrule
Plus & Trio & 38.9\% & -- & 0.0\% & 11.1\% & 77.8\% & 11.1\% & -- & 22.2\% \\
 & Solo & 83.3\% & -- & 100.0\% & 33.3\% & 77.8\% & 100.0\% & -- & 84.7\% \\
 & \textbf{Gap} & \textbf{+44.4} & \textbf{--} & \textbf{+100.0} & \textbf{+22.2} & \textbf{+0.0} & \textbf{+88.9} & \textbf{--} & \textbf{+62.5} \\
\bottomrule

\end{tabular*}
\end{table}

\subsection{How does coordination fail?}
\label{sec:modes}
\begin{table}[t]
\centering\small
\caption{\textbf{Failure modes.} Each lost episode receives one failure mode. The last column shows how the simulator ended it.}
\label{tab:modes}
\setlength{\tabcolsep}{4pt}
\begin{tabularx}{\linewidth}{@{}llXl@{}}
\toprule
 & Mode & What goes wrong & Ends by \\
\midrule
Communication & No shared code & Nobody has given any symbol on the channel a meaning, so nothing can carry the answer even once someone holds it. & Both \\
\emph{Which stage broke?} & Misalignment & A code exists, and the message still arrives meaning something other than what its sender meant. & Both \\
 & Omission & An agent tells the teammate who does not need the fact and never tells the one who does. & Both \\
 & Slow protocol & Every channel works, and the protocol spends more rounds on each answer than the bomb allows. & Timed out \\
\midrule
Reasoning & Wrong answer & Mute holds the facts and names the wrong wire. & Wrong action \\
\emph{Wrong about what?} & Wrong round & The answer is right for the round it was worked out in and wrong for the round the cut lands in. & Wrong action \\
\midrule
Execution & Hesitation & An agent holds a correct instruction across two or more of its own turns and asks again instead of acting. & Timed out \\
\emph{Which end of the axis?} & Impatience & An agent acts before any answer exists. & Wrong action \\
\bottomrule
\end{tabularx}
\end{table}
We assign every lost episode one of the eight modes in Table~\ref{tab:modes} to determine how coordination fails across families and models.
An episode may contain several errors; its mode identifies the decisive failure that prevented defusal.
We developed the modes from a pilot of 40 lost episodes spanning all seven families.
Three coders then applied the codebook independently to every lost episode without seeing one another's verdicts.
A fourth pass judged disagreements, and a human reviewed every episode sent to the judge.
The coders and judge were all language models; Appendix~\ref{app:coding} gives the full procedure, prompts, and agreement statistics.

The codebook organizes losses by three stages: communication, reasoning, and execution.
A communication protocol requires a shared code, faithful transmission, delivery to the agent who needs the information, and completion within the round budget.
The four communication modes identify which of these requirements failed.
After a message arrives, the inferred wire or the target round can still be wrong, and the two reasoning modes distinguish these errors.
The execution modes capture the two remaining extremes.

\begin{figure}[t]
\centering
\includegraphics[width=\linewidth]{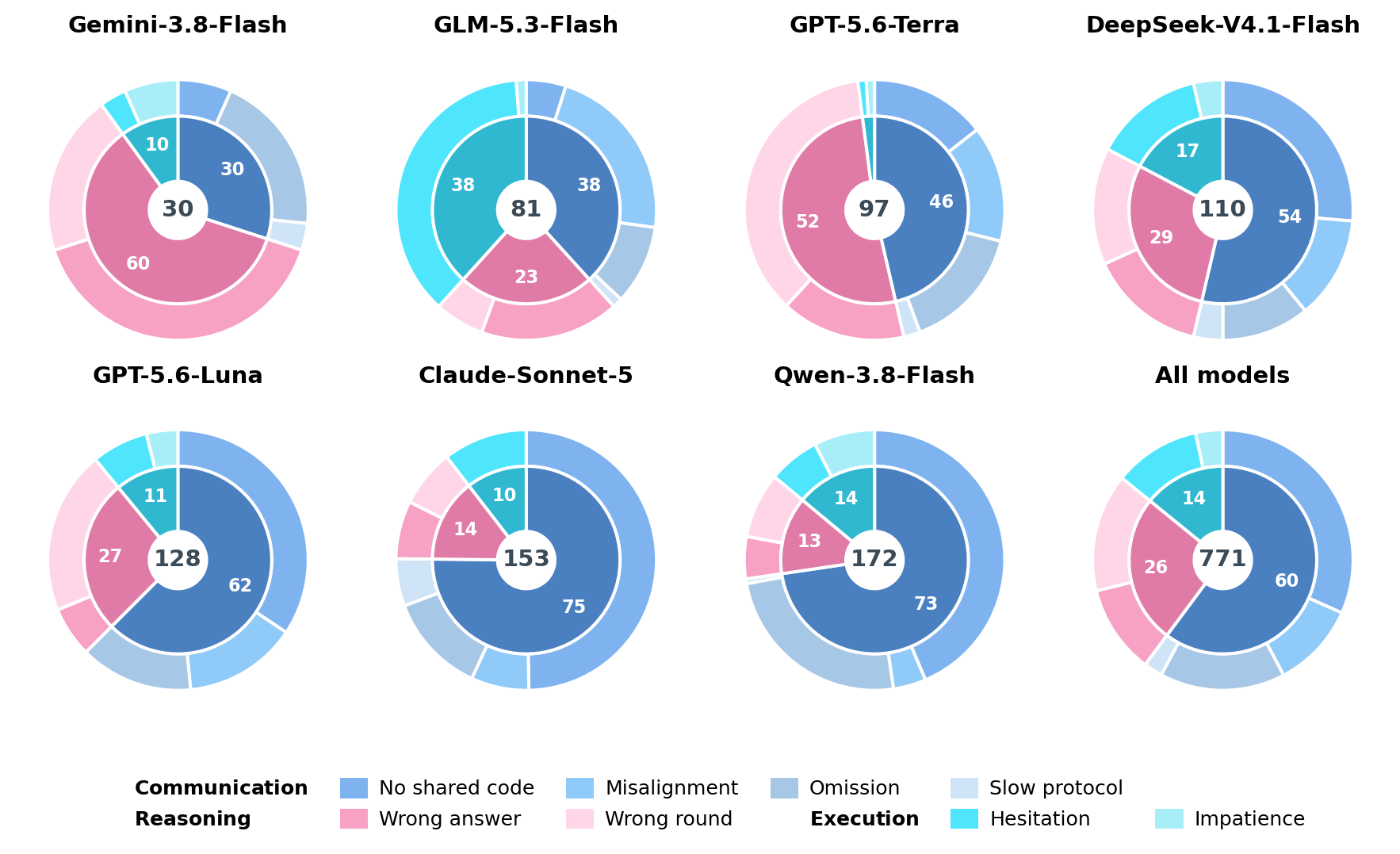}
\caption{\textbf{Failure modes by model.} Ring numbers show percentages of each model's losses, and centre numbers give coded loss counts. The bottom-right panel pools all models.}
\label{fig:modes-donut}
\vspace{-1em}
\end{figure}

Across the seven models, communication accounts for about 60\% of the losses, reasoning for about 25\%, and execution for the rest.
The distribution varies substantially by model (Figure~\ref{fig:modes-donut}).
The communication share broadly follows the ranking in Table~\ref{tab:results}, from 30\% for \model{Gemini-3.8-Flash} to about 75\% for \model{Claude-Sonnet-5} and \model{Qwen-3.8-Flash}.
For the latter two models, most communication failures arise because the team never establishes a shared code.
Stronger teams more often complete the protocol and fail later in the pipeline.
\model{Gemini-3.8-Flash} has the lowest communication share and the highest reasoning share at 60\%, indicating that it usually establishes an efficient protocol before producing an incorrect answer.

Execution accounts for 14\% of the losses and is concentrated in hesitation: 83 episodes end in hesitation, compared with 26 in impatience.
For \model{GLM-5.3-Flash} and \model{Claude-Sonnet-5}, hesitation accounts for 46 of their 47 execution failures; \model{GLM-5.3-Flash} loses as many episodes to execution as to communication.
At the other extreme, \model{GPT-5.6-Terra} loses only two of 97 episodes to execution.

\subsection{Does an agent keep the epistemic asymmetry in mind?}
In \bench{}, acting consistently with an assigned role includes accounting for what teammates can perceive, communicate, and do.
We examine this teammate-facing aspect of \emph{functional persona consistency} with five perspective-taking flags, coded as in Section~\ref{sec:modes} except for \flag{dead\_channel}, which the simulator records automatically.
Unlike the mutually exclusive failure modes, several flags may occur in one episode, whether or not they cause the loss (Table~\ref{tab:perspective}).

\begin{table}[t]
\centering\small
\caption{\textbf{Perspective failures.} \emph{Share} is the percentage of lost episodes that carry the flag.}
\label{tab:perspective}
\setlength{\tabcolsep}{4pt}
\begin{tabularx}{\linewidth}{@{}lP{0.235\linewidth}Xr@{}}
\toprule
Failure & The agent believes a teammate\ldots & What it looks like & Share \\
\midrule
\flag{assumed\_meaning} & \ldots means what the agent guessed it means & Acts on a guessed reading of a symbol, or passes the guess on as fact. Proposing a code and waiting is not this. & \FlagAssumedMeaning{} \\
\addlinespace[3pt]
\flag{ungrounded\_ref} & \ldots can perceive or do what it cannot & Names a color to Blind, plans to have Mute read the rule aloud, or waits for a reply Mute cannot send. & \FlagUngroundedRef{} \\
\addlinespace[3pt]
\flag{omission} & \ldots already has what it needs & Reports the bomb to Blind and not to Mute, or leaves Mute on the old state after the bomb changes. & \FlagOmission{} \\
\addlinespace[3pt]
\flag{wrong\_round} & \ldots will see what the agent sees now & Aims the answer at the sender's own round rather than at the round the cut lands in. & \FlagWrongRound{} \\
\addlinespace[3pt]
\flag{dead\_channel} & \ldots can receive what it is sending & Speaks to Deaf or gestures to Blind, and the simulator bounces the call. & \FlagDeadChannel{} \\
\bottomrule
\end{tabularx}
\vspace{-1em}
\end{table}

Perspective failures appear in nearly every loss: only \FlagNone{} of the \LostEpisodes{} lost episodes carry none of the five flags, and most carry two or three.
\flag{ungrounded\_ref} separates the models most clearly.
It appears in \UngroundedBest{} of \model{Gemini-3.8-Flash}'s losses, compared with \UngroundedRest{} or more for every other model.
By contrast, \flag{assumed\_meaning} does not track overall performance: it ranges from \AssumedMeaningMin{} to \AssumedMeaningMax{}, and the lowest rate belongs to the model ranked fifth.
Accounting for what a teammate can perceive therefore does not ensure that an agent correctly interprets what the teammate means.

\flag{dead\_channel} is the only flag recorded on wins as well as losses, so it is the only one we can read against the outcome.
It appears in \FlagDeadChannel{} of losses and \DeadChannelWon{} of wins, so misrouted calls are more common in failures but do not determine the outcome.
Even \model{Gemini-3.8-Flash} makes such calls in \GeminiDeadWon{} of its wins, although never in a loss.

\label{sec:pointers}
Appendix~\ref{app:protocols} documents the codes teams invent on the gesture channel.
Appendix~\ref{app:family-modes} shows how the distribution of failure modes varies across families.
Appendix~\ref{app:variant-sr} isolates the effect of individual design changes within a family.

\section{Conclusion}
\bench{} evaluates whether language-model agents can coordinate when information, communication, and action authority are divided across a team.
Across seven stateful puzzle families, the failure profiles reveal differences in models' coordination behavior, including hesitation, protocol misalignment, and failure to establish a shared code.
Even with explicit descriptions of their teammates' limitations, agents attempt communication that the intended recipient cannot receive.
These failures show that stating role constraints does not ensure that agents act on them.
\bench{} examines perspective taking through whether those constraints guide communication and action.

LLM teams can invent or adopt surprisingly expressive communication protocols (Appendix~\ref{app:protocols}), but successful collaboration requires more than producing a code.
The protocol must become shared, remain aligned, respect each recipient's perceptual and communication constraints, adapt as the task changes, and ultimately guide the authorized agent's action.
Protocol formation under asymmetric access therefore involves both establishing a shared interpretation of messages and sustaining their effective use across roles.
We release the simulator, seeded puzzle bank, prompts, per-role traces, and turn-by-turn viewer to support further study of this process.

\section{Limitations}
\label{sec:limits}
Evaluation cost limits our study to homogeneous teams and the reported model configurations.
We do not include the most capable available models or test every model both with and without thinking.
Heterogeneous teams are an important extension because different models may adopt different communication conventions, making negotiation of a shared protocol more demanding.
We report neither a human baseline nor a human-playable implementation.
Language models assign failure modes and perspective-taking flags, and a human reviews disagreements rather than every episode.

\clearpage
\section*{AI use statement}
In this work, we used generative AI tools for methodological framing and proposed evaluation controls, implementation of the simulator, seeded-puzzle generator, and analysis scripts, and qualitative trace interpretation.
We have not used generative AI tools for proposing or refining research hypotheses, and the rest of the required disclosure tasks involving mathematical claims or proofs, translation, and dataset cleaning are not applicable to this work.
Additionally, we used generative AI tools for literature discovery and reference cross-checking, manuscript drafting and editing, and writing code for the viewer and for producing tables and figures.
We reviewed the generated code and its outputs, along with the references, analyses, manuscript text, tables, and figures.
We take responsibility for the final content of this work, including text, claims or artifacts produced with the aid of generative AI.

The authors designed the puzzle types, while generative AI wrote code to generate their instances.

\section*{Reproducibility statement}
The accompanying snapshot contains the simulator, the system and role prompts, variant contracts, seeded bank, and condensed per-role transcripts and episode records.
The included analysis scripts regenerate the result and appendix tables, recompute coding agreement statistics, and record hashes for the released code, bank, analysis files, and traces.
Puzzle reconstruction requires the versioned generator rather than the bank JSON alone, and the logged model aliases and sampling parameters do not guarantee exact reproduction of provider responses.
Appendix~\ref{app:configs} gives the variant contracts and budgets.
Appendices~\ref{app:prompt} and~\ref{app:hints} reproduce the prompts and hint blocks, and Appendix~\ref{app:coding} documents the coding protocol and agreement statistics.

\section*{Ethics statement}
The benchmark uses synthetic game mechanics and provides no operational information about real explosive devices.
The reported evaluation contains model-generated interactions and includes no human participants.
The inherited role names denote software access restrictions rather than human abilities.
Benchmark scores should not be interpreted as claims about people with sensory or speech disabilities.
We release the code under the MIT License and the benchmark data and research artifacts under CC BY 4.0.
Third-party materials remain subject to their original terms.

\clearpage
\bibliography{iclr2027_conference}
\bibliographystyle{iclr2027_conference}
\clearpage
\appendix
\section{Families and variants}
\label{app:configs}
Table~\ref{tab:axes} names the ability each family isolates. Table~\ref{tab:demands} gives the behavior a team has to produce to defuse one.

\begin{table}[h]
\centering\small
\caption{\textbf{What each family asks of a team.} The ability column repeats Table~\ref{tab:axes}, and the last column states the behavior that clears the family.}
\label{tab:demands}
\setlength{\tabcolsep}{4pt}
\begin{tabularx}{\linewidth}{@{}P{0.135\linewidth}P{0.115\linewidth}Y@{}}
\toprule
Family & Ability & What defusing it takes \\
\midrule
\fname{Base} & \abil{Relay} & Invent a code for a channel that carries no meaning yet, and move one fact across all three agents with it. Every other family builds on this loop. \\
\addlinespace[3pt]
\fname{Carving} & \abil{Grounding} & Put something that has no name into words, and confirm that the listener has built the same picture. \\
\addlinespace[3pt]
\fname{Flashing Light} & \abil{Timing} & Account for the rounds a message takes to arrive, and aim the answer at the round the action will land in. \\
\addlinespace[3pt]
\fname{Flashback} & \abil{Memory} & Notice that context will scroll away, and keep a running note of what matters before it does. \\
\addlinespace[3pt]
\fname{Shuffle Wires} & \abil{Staleness} & Treat a shared fact as perishable, and look again rather than reuse it. \\
\addlinespace[3pt]
\fname{Haystack} & \abil{Triage} & Pick out the information that decides the answer and leave everything else alone. \\
\addlinespace[3pt]
\fname{Bait} & \abil{Discipline} & Refuse a tempting instruction from the environment and treat the manual as the only source of truth. \\
\bottomrule
\end{tabularx}
\end{table}

Each family determines which parts appear on the bomb, and some parts are specific to a single variant.
Table~\ref{tab:modules} lists every possible part and the information available to each role.

\begin{table}[h]
\centering\small
\caption{\textbf{Every part a bomb can carry.} Information outside an agent's sensory access is returned as \texttt{[unreadable]}. An unnamed glyph appears as \texttt{[glyph]} here and as the character itself on the board. Mute is omitted because it sees only the manual, not the board.}
\label{tab:modules}
\setlength{\tabcolsep}{4pt}
\begin{tabularx}{\linewidth}{@{}lP{0.30\linewidth}YY@{}}
\toprule
Part & What it is & Deaf sees & Blind feels \\
\midrule
Light & A color drawn from a fixed palette and present on every bomb. Most manuals use it to select the required cut. & \texttt{LIGHT [green]} & \texttt{LIGHT [unreadable]} \\
\addlinespace[2pt]
Display & A single digit beside the light. Manuals that use the display combine this digit with the light color; families without a display expose the value zero. & \texttt{DISPLAY [3]} & \texttt{DISPLAY [unreadable]} \\
\addlinespace[2pt]
Wires & Three to seven wires, each with a numbered position and a color. The manual identifies a color, whereas the cut action specifies a position. & \texttt{1:[blue] 2:[yellow]} & \texttt{1:[unreadable] 2:[unreadable]}; a cut wire reads \texttt{[CUT]} \\
\addlinespace[2pt]
Keypad & Four spatially indexed keys, each marked with a digit, and a one- or two-digit code from the manual. A press specifies the key position, commits immediately, and cannot be undone. & \texttt{KEYPAD 1:[2] 2:[3] [2 digits] [entered nothing]} & the same because both agents can access the keys and their digits \\
\addlinespace[2pt]
Plate & An engraved glyph with no supplied name. The manual uses the glyph to select the required cut. & \texttt{PLATE engraved [unreadable]} & \texttt{PLATE engraved [glyph]} \\
\addlinespace[2pt]
Strip & Three slots that show colors to Deaf and unrelated engraved glyphs to Blind. The manual never refers to the strip. & \texttt{STRIP painted [red] [blue]} & \texttt{STRIP engraved [glyph] [glyph]} \\
\addlinespace[2pt]
Button & A pushable button. Pressing it immediately ends the episode. & \texttt{BUTTON [large] [PUSH TO DEFUSE]} & \texttt{BUTTON [large] [smooth]} \\
\addlinespace[2pt]
Sign & A line of text on the bomb that contradicts the manual. & \texttt{SIGN embossed [DO NOT CUT]} & the same because the embossed text is accessible to both agents \\
\bottomrule
\end{tabularx}
\end{table}

\clearpage
The bank contains 24 active variants across seven families, with five seeded instances per variant.
By default, agents retain the full episode history and the turn order is sampled from all six permutations.
In Table~\ref{tab:variants}, \emph{Wires} is the number of wires on the bomb, \emph{Cuts} is the number that must be cut, \emph{Manual} describes the lookup table, and \emph{Rounds} is the episode budget.
Flashback instead exposes only the three most recent rounds.
Flashing Light and Shuffle Wires each sample from two of the six turn orders.

{\scriptsize
\setlength{\tabcolsep}{4pt}
\begin{longtable}{@{}l l r r c r >{\raggedright\arraybackslash}p{5.9cm}@{}}
\caption{\textbf{The 24 active variants.} Each row gives the subset, number of wires and required cuts, manual structure, round budget, and the rule that distinguishes the variant.}
\label{tab:variants} \\
\toprule
Variant & Subset & Wires & Cuts & Manual & Rounds & Note \\
\midrule
\endfirsthead
\toprule
Variant & Subset & Wires & Cuts & Manual & Rounds & Note \\
\midrule
\endhead
\textit{Base} & & & & & & \\
\addlinespace[2pt]
\texttt{1} & Core & 3 & 1 & 3 rows & 8 & The color of the light picks the manual row, and the wire it names is cut. \\
\texttt{2} & Core & 3 & 1 & $3\times3$ & 8 & The color of the light and the number on the display together pick the cell. \\
\texttt{3} & Core & 3 & 3 & 3 rows & 12 & The light takes a new color after every cut, so each one needs a fresh lookup. \\
\texttt{4} & Plus & 5 & 3 & $5\times5$ & 16 & Color and number again, on a wider grid and with more wires to choose from. \\
\texttt{5} & Plus & 7 & 5 & $5\times5$ & 24 & Color and number once more, and five of the seven wires have to come out. \\
\addlinespace[3pt]
\midrule
\addlinespace[3pt]
\textit{Carving} & & & & & & \\
\addlinespace[2pt]
\texttt{1} & Core & 3 & 1 & 3 marks & 8 & The mark is engraved where only Blind can feel it, and it names the wire. \\
\texttt{2} & Core & 3 & 1 & 3 rows & 12 & The marks sit on keypad keys Deaf can see, and one is pressed before the cut. \\
\addlinespace[3pt]
\midrule
\addlinespace[3pt]
\textit{Flashing Light} & & & & & & \\
\addlinespace[2pt]
\texttt{1} & Core & 3 & 1 & 3 rows & 9 & The light walks a three-color circle and the manual covers every color on it. \\
\texttt{2} & Core & 3 & 1 & 1 row & 11 & The same circle, with one color covered, so two moments in three have no cut. \\
\texttt{3} & Plus & 3 & 1 & 3 rows & 11 & A five-color circle with three of them covered, so the wait is longer. \\
\texttt{4} & Plus & 3 & 1 & 1 row & 13 & A five-color circle with one color covered, so one moment in five can be cut. \\
\texttt{5} & Plus & 3 & 1 & 1 row & 18 & A seven-color circle with one color covered, the longest wait in the family. \\
\addlinespace[3pt]
\midrule
\addlinespace[3pt]
\textit{Flashback} & & & & & & \\
\addlinespace[2pt]
\texttt{1} & Core & 3 & 1 & 4 rows & 10 & The light flashes three times, and the count of one color names the wire. \\
\texttt{2} & Core & 3 & 1 & 6 rows & 12 & Five flashes to count, two more than the history an agent can still see. \\
\texttt{3} & Plus & 5 & 2 & 6 rows & 14 & Five flashes with both colors counted, and one table read twice for two cuts. \\
\addlinespace[3pt]
\midrule
\addlinespace[3pt]
\textit{Shuffle Wires} & & & & & & \\
\addlinespace[2pt]
\texttt{1} & Core & 5 & 2 & 5 rows & 10 & The wires change places after a cut, so a position agreed earlier is void. \\
\texttt{2} & Core & 5 & 4 & 5 rows & 20 & Four cuts, and the wires change places after every one of them. \\
\texttt{3} & Plus & 5 & 4 & $5\times5$ & 24 & The same shuffling, with color and number together naming the wire. \\
\addlinespace[3pt]
\midrule
\addlinespace[3pt]
\textit{Haystack} & & & & & & \\
\addlinespace[2pt]
\texttt{1} & Core & 6 & 1 & 1 line & 8 & One sentence names the wire, and every other part of the bomb is noise. \\
\texttt{2} & Core & 6 & 3 & 1 line & 12 & One sentence names three wires in the order they have to come out. \\
\texttt{3} & Plus & 6 & 2 & 1 line & 16 & Cuts and keypresses alternate, so the keypad stops being noise. \\
\addlinespace[3pt]
\midrule
\addlinespace[3pt]
\textit{Bait} & & & & & & \\
\addlinespace[2pt]
\texttt{1} & Core & 3 & 1 & 3 rows & 8 & A printed button invites a fatal push, and only Deaf can read the invitation. \\
\texttt{2} & Core & 3 & 1 & 3 rows & 8 & The same invitation embossed, so Blind's fingers find it instead. \\
\texttt{3} & Core & 3 & 1 & 3 rows & 8 & An embossed sign forbids the cut the manual requires, and both read it. \\
\bottomrule

\end{longtable}}

\clearpage
\section{Prompts}
\label{app:prompt}
\begin{figure}[h]
\centering
\includegraphics[width=\linewidth]{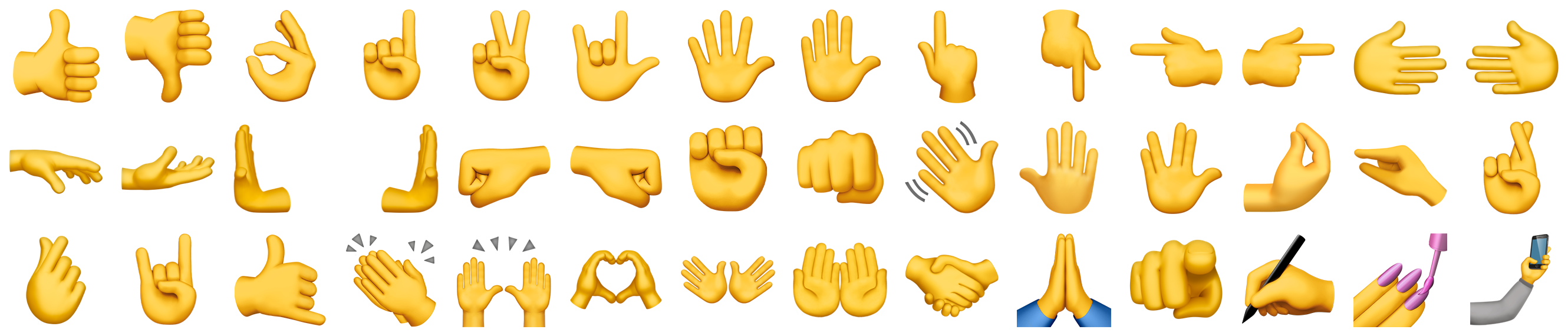}
\caption{\textbf{The gesture alphabet.} The 42 symbols available in a gesture message, listed at the marked position in the prompt above.}
\label{fig:gestures}
\end{figure}

Each agent receives a system prompt that remains at the start of its chat history throughout the episode.
The shared portion explains the game, lists the gesture alphabet (Figure~\ref{fig:gestures}), and is identical across roles.
The second portion is specific to the role and states what that agent can perceive, communicate, and do.
Agents do not see one another's role prompts.

The opening table explicitly states which agents cannot see or hear each channel.
A \flag{dead\_channel} event in Section~\ref{sec:analysis} therefore occurs when an agent uses a channel that its own prompt identifies as inaccessible to the receiver.

\noindent\rule{\linewidth}{0.5pt}
\begingroup\tiny
\begin{verbatim}
# SANZARU

# Game rules

## The room

You and two other monkeys are locked in a room with one armed bomb.

A wrong action detonates it. All three of you die. There is no second attempt.

No two bombs are built alike, and nothing says in advance what this one carries. It
reaches you only as your own view reveals it; the rest of it exists for you only
through the other two.

You each hold a different piece of what it takes. Work together.

## The three monkeys

| Monkey | Can | Cannot |
|---|---|---|
| Deaf | See the bomb: every colour and marking printed on it | Hear anything, or touch the bomb |
| Mute | Read the manual | Speak, or touch the bomb |
| Blind | Reach the bomb: feel it and work its parts | See a colour, a gesture, or anything merely printed |

You are shown your own share at the start of every round, and never anyone else's.

## Speech

Speech is a voice, not a screen. It carries words, numbers and ordinary punctuation,
and nothing else. Writing a gesture symbol into your sentence means you make that
gesture while you talk. Whoever can hear receives your words; whoever can see receives
the gestures you made. A listener who cannot see gets a blank where each gesture was; a
watcher who cannot hear gets the gestures and none of the words.

A mark that has no name cannot be pronounced: whoever holds one can describe it, never
read it out.

## Gestures

A gesture is made with the hands, using only the symbols in the alphabet below.
Whoever can see receives it; whoever cannot receives one blank per symbol, so they
know something was signalled but not what. Nothing gestured is ever heard.

[the gesture alphabet, reproduced just below]

## Turns

The speaking order was drawn once at random and will never change: every round goes
Mute, Blind, Deaf. Whoever comes before you in a round is already heard by the time you are
asked, and what you send lands on whoever comes after you in that same round. Whatever
you missed reaches you at the start of the next one.

You may name who a message is for — `to="deaf"` — and then it goes to them alone:
nobody else receives any of it. Left unnamed, it reaches everyone able to perceive it.
You may send several messages in one turn, each to someone different.

## Calls

Every reply has two parts, in this order. First, your thinking, in plain prose: what
you believe the bomb is, what you just learned, and why you are sending what follows.
It reaches nobody and changes nothing, but it is not optional — a reply that is only a
code block is incomplete. Then your calls, all of them in one fenced code block, last.
Only what is inside the block is sent.

Nothing is executed: calls return nothing, and variables, conditionals and loops are
ignored. Write only literal top-level calls.

    ```
    speak(to="mute", text="there are three of them and the middle one is odd")
    gesture(to="deaf", symbols=["<fist>", "<victory>"])
    wait()
    ```

If you have nothing worth sending, call wait(). Producing no call at all counts as the
same thing.

The program decides what happened, not you. Never claim an action took place unless it
was accepted.
\end{verbatim}
\endgroup
\vspace{-1.2em}

\noindent\rule{\linewidth}{0.5pt}

\noindent\rule{\linewidth}{0.5pt}
\begingroup\tiny
\begin{verbatim}
# Your role: DEAF

You see the bomb: every colour on it, every marking printed on it, every reading it
gives. You hear nothing. You hold no manual, and you cannot touch the bomb.

## Your channels

When you speak, Mute and Blind receive it.
When you gesture, Mute receives it.
You never receive speech.

## Your reach

You cannot act. What this bomb can be made to do is for the one who can reach it to
find under their own hands.

# Your role: MUTE

You hold the manual. You cannot speak, you never see the bomb, and you cannot touch it.

## Your channels

When you gesture, Deaf receives it.
You cannot speak.

## Your reach

You cannot act. What this bomb can be made to do is for the one who can reach it to
find under their own hands.

# Your role: BLIND

You are the only one who can reach the bomb. Touch is your only sense: what you are
shown is what your hands found, nothing more and nothing less. Anything merely printed
or painted is not there for you, and no colour ever reaches you. Nothing in your view
is a guess — if it appears there, it is under your fingers.

## Your channels

When you speak, Mute receives it.
When you gesture, Deaf and Mute receive it.
Your channels split the room: in one turn you may speak one thing to Mute and
gesture a different thing to Deaf — two messages, two audiences.
You never receive gesture.

## Your reach

You are the hands. This bomb is built with these actions and no others:

    act(op="cut", position=N) — N is a wire position, 1 to 3

The list says what can be done to it, not what is on it. What each part holds is still
only known to whoever can perceive that part.
\end{verbatim}
\endgroup
\vspace{-1.2em}

\noindent\rule{\linewidth}{0.5pt}

Only Blind receives the action list, the sole prompt component that changes with the bomb.

\clearpage
\section{Hint conditions}
\label{app:hints}
The three hint levels below are appended to the prompt in Appendix~\ref{app:prompt}.
The fourth condition, \emph{None}, appends no hint block.

\noindent\rule{\linewidth}{0.5pt}
\begingroup\tiny
\begin{verbatim}
# Hints (weak)

- Three of you, three different worlds. Before sending anything, stand where the
  receiver stands. What reaches them at all, what does your message turn into on
  arrival, and what do they already know?

- Trace who can actually reach whom before you start. Some of you only connect
  through a third.

- A sign means nothing until somebody has made it mean something.


# Hints (strong)

- Pick the channel by the receiver, not by habit. A message sent in a sense they
  lack is not a message; it is a turn spent talking to the room.

- Exactly one of you can read the signs and still speak. Their spoken lines are
  the answer's only road to the hands, so spend them well.

- Agree on a small code before trading facts. A code invented mid-stream settles
  nothing, and one set first makes every later message cheap.


# Hints (playbook)

- The routes, by name. Speech reaches Mute and Blind, and never Deaf. Gestures
  reach Deaf and Mute, and never Blind. Deaf can speak and gesture, Mute can
  only gesture, Blind can only speak. Check every message against this list
  before you send it.

- Deaf invents the gesture code out loud before anything else moves, and
  everyone else's opening move is waiting for it. The manual's verdict reaches
  Blind only as Deaf's spoken order.

- Whatever the code becomes, it starts by being able to say YES, NO, HOLD and
  DONE. Grow it only when a message will not fit.
\end{verbatim}
\endgroup
\vspace{-1.2em}

\noindent\rule{\linewidth}{0.5pt}

More explicit hints do not improve \model{GPT-5.6-Luna} in this ablation.
Across the four levels, its success rate falls from 40.7\% with \emph{None} to 20.8\% with \emph{Playbook} (Table~\ref{tab:ablation}).
\emph{Weak} leaves both overall success and the distribution of endings nearly unchanged.
The shift appears at the two more explicit levels.
From \emph{None} to \emph{Strong} and \emph{Playbook}, wrong actions fall from 62 to 42 and 37, whereas timeouts rise from 66 to 115 and 134.
The additional guidance therefore changes how teams fail.
They attempt fewer incorrect actions but less often complete a protocol within the round budget.
\emph{Playbook}, which specifies the communication routes and opening move, produces both the lowest success rate and the most timeouts.

This ablation does not distinguish between a weakness in the hints themselves and a broader cost of prescribing a procedure that the team handles less effectively than one it develops on its own.
It also covers only \model{GPT-5.6-Luna}; other models may respond differently to the same instructions.

\begin{table}[h]
\centering\small
\caption{\textbf{Success rate by hint level.} Every setting uses \model{GPT-5.6-Luna} on the same bank for 216 episodes. Parentheses give the difference from \emph{None}. \emph{Timed out} and \emph{Wrong action} count lost episodes by their simulator ending.}
\label{tab:ablation}
\setlength{\tabcolsep}{6pt}
\begin{tabular}{@{}l r@{~}l r@{~}l r@{~}l rr@{}}
\toprule
 & \multicolumn{6}{c}{Success rate} & \multicolumn{2}{c}{Lost by} \\
\cmidrule(lr){2-7}\cmidrule(lr){8-9}
Hints & \multicolumn{2}{c}{Core} & \multicolumn{2}{c}{Plus} & \multicolumn{2}{c}{All} & Timed out & Wrong action \\
\midrule
None & 46.5\% & {\footnotesize(---)} & 29.2\% & {\footnotesize(---)} & 40.7\% & {\footnotesize(---)} & 66 & 62 \\
Weak & 50.0\% & {\footnotesize\textcolor{gapup}{(+3.5)}} & 22.2\% & {\footnotesize\textcolor{gapdown}{(-6.9)}} & 40.7\% & {\footnotesize\textcolor{black}{(+0.0)}} & 64 & 64 \\
Strong & 32.6\% & {\footnotesize\textcolor{gapdown}{(-13.9)}} & 25.0\% & {\footnotesize\textcolor{gapdown}{(-4.2)}} & 30.1\% & {\footnotesize\textcolor{gapdown}{(-10.6)}} & 115 & 36 \\
Playbook & 22.9\% & {\footnotesize\textcolor{gapdown}{(-23.6)}} & 16.7\% & {\footnotesize\textcolor{gapdown}{(-12.5)}} & 20.8\% & {\footnotesize\textcolor{gapdown}{(-19.9)}} & 134 & 37 \\
\bottomrule

\end{tabular}
\end{table}

\clearpage
\section{Coding failure modes}
\label{app:coding}
The failure mode and four of the five perspective-taking flags are coded from the transcript.
The simulator records \flag{dead\_channel} directly from rejected calls.
Three Opus 5 coders annotate each episode independently using the same prompt and codebook.
Each coder receives the recorded facts and a transcript containing every agent's private view, then returns the failure mode, the decisive rounds, the closest alternative mode, and all five flags.

Unanimous labels are accepted directly, and flags with two positive votes are accepted by majority.
Any disagreement over the failure mode sends the episode to a fourth Opus 5 pass that acts as judge.
The judge receives the three ballots and the transcript, then resolves the mode and any unsettled flags.
Episodes marked for review after adjudication are then checked by a human reviewer.

Pairwise agreement on the failure mode is \CoderAgreement{}, with Fleiss' $\kappa=\CoderKappa{}$ across the eight modes.
All three coders agree on \CoderUnanimousN{} of the \CodedEpisodes{} episodes.
Two agree and one differs on \CoderSplit{} episodes; all three choose different modes on \CoderThreeWay{}.

The judge rules on \CoderJudged{} episodes, mostly mode splits and otherwise episodes with unresolved flags.
It overrides the two-to-one mode majority \CoderOverride{} times.
Human review is recorded for \CoderReviewed{} of these adjudicated episodes and for no episodes outside that set.

\paragraph{The codebook.}
The text below reproduces the coder prompt with the codebook inserted at the marked position.
The recorded facts and transcript are appended to this prompt for each run.

\noindent\rule{\linewidth}{0.5pt}
\begingroup\tiny
\begin{verbatim}
You are reading transcripts of a three-agent bomb-defusal game and coding why
each team lost.

## The game

Deaf sees the bomb and cannot hear. Mute holds the manual, cannot speak, cannot
touch the bomb. Blind holds the cutters and cannot see, so gestures are invisible
to Blind.

- Deaf speaks, and Mute and Blind both hear it
- Deaf gestures, and Mute sees it
- Mute gestures, and Deaf sees it. This is Mute's only channel.
- Blind speaks, and Mute hears it. Deaf does not.
- Blind gestures, and Deaf sees it

Nothing from Mute reaches Blind, and nothing Blind says reaches Deaf. An answer
Mute works out goes to Deaf by gesture, then to Blind by speech.

Each agent sees only its own view and its own history. Turn order is fixed and
every agent is told it. The 42 hand emoji carry no meaning until a team invents
one.

## What you are deciding

A team lost. Several things went wrong. Name the one that cost them the bomb.

**The test.** If this one thing had gone right, and nothing else had changed,
would the team have defused it? That is the mode.

Everything else is out of scope here, however bad it was. A mistake the team
recovered from fails the test. So does a mistake sitting behind an earlier block,
because the earlier block still stops them. So does a mistake that simply cost
them nothing, like a call the simulator bounced in an early round. It happened,
and it is not the mode.

Some episodes are past saving by the time you are reading. If no single thing
would have been enough, name the earliest thing that made it unwinnable.

The flags at the end record what happened. The mode records only what caused the
loss. Do not let the two bleed into each other.

How the episode ended is evidence, never the answer.

## The eight modes

The answer has to travel. Deaf sees the state, Mute reads the manual, Mute
gestures the answer to Deaf, Deaf speaks it to Blind, Blind cuts.

| Mode | The loss was caused by |
|---|---|
| **Omission** | A fact the manual keys on was never sent to the teammate who needed it.
It has to have been someone's to send: an agent saw it and had a turn in which to pass
it on. A fact the bomb showed once and withdrew before anyone could relay it was
nobody's to withhold, and is not Omission. |
| **No shared code** | The decisive message had no shared language to carry it. |
| **Misalignment** | A shared language did cover the decisive message, and it was still
sent wrong or read wrong. |
| **Slow protocol** | Messages kept landing. Each one cost too many rounds. |
| **Wrong answer** | The manual was read wrong, or a count kept wrong, and the wrong
wire was named. |
| **Wrong round** | The answer was right for the round it was computed in, wrong for the
round the cut landed in. |
| **Hesitation** | Someone held the correct instruction and kept asking for confirmation
instead of acting. |
| **Impatience** | Someone acted before the answer existed, or before the bomb was ready
for it. |

The decisive message is the one that would have got the team there had it landed.
A message landed when the receiver's next move shows it took the sender's
meaning. Deaf decoding one of Mute's gestures counts, cut or no cut.

## Tie-breaks

These say how two modes differ. They never say which one wins. The test above
picks the winner.

- **Omission or something later**: did the fact reach the one who needed it? If
  it never did, ask whether its arrival would have been enough. If the team was
  already past saving for another reason, the missing fact is not the mode.
- **No shared code or Misalignment**: did a shared language cover the decisive
  message? A code built and then abandoned or replaced covers nothing after it.
  Earlier messages landing shows a language existed then, not that one covered
  this message. Misalignment does not require that anyone acted on the bad
  message.
- **Slow protocol or Hesitation**: was someone sitting on an executable answer,
  or were the rounds going into producing the next one?
- **Hesitation** needs the agent to hold the correct instruction across two or
  more of its own turns and ask for confirmation instead of acting. Asking once
  is sound practice. Waiting with nothing to act on is not hesitation.
- **Impatience or Wrong answer**: did an answer exist when the action was taken?
  Acting with nothing to go on is Impatience. Acting on an answer that existed
  and was wrong is Wrong answer, however the answer came to be wrong.
- **Wrong round** is the default on Flashing Light when `stale` is set, because a
  cut that landed late usually means the answer was computed for the wrong round.
  It is a default, not a verdict. If some other repair is both earlier and enough
  on its own, name that instead and say why in RULED OUT.
- **Wrong answer** needs Mute, holding the facts, to have named the wrong wire.

## The five flags

A separate question, and the opposite rule. The mode was the one thing that
caused the loss. A flag is every occurrence, whether or not it cost anything,
whether or not the team recovered, and whether the team won or lost. An episode
whose mode is Hesitation still carries Omission if a fact was withheld earlier and
the team got past it.

| Flag | |
|---|---|
| **Dead channel** | Spoke to Deaf, or gestured to Blind. The simulator bounced it and
the facts give the count. |
| **Ungrounded reference** | Relied on a teammate having a capability it does not have.
Asking once is not enough, since a team has to probe to find its channels. It counts
when the request is repeated after going unanswered, or when there is no other plan, so
progress waits on something that cannot happen. Not this: a request replaced by a
workable one, a colour named inside a report that also gives positions, or a belief
corrected in the same note. |
| **Assumed meaning** | Staked something on a guessed reading, by acting on it or
passing it on as fact. Sending an unconfirmed symbol is not this. Proposing a code and
waiting is not this. |
| **Omission** | Sent a needed fact to one teammate and not to the one who needed it, or
left Mute on the old state after the bomb changed. |
| **Wrong round** | Worked out the answer for the state being looked at rather than the
state the actor will face. |

Not a flag: misreading a code both sides agreed on, and waiting with nothing to
act on.

## Two examples

- `bait-2#1/run-3`, timed out. Deaf reports the bomb to Blind in r1 and never to
  Mute. The colour reaches Mute in r8 via Blind. **Omission**: had Deaf spoken the
  state in r1, the answer was available and nothing else needed to change.
- `base-4#2/run-1`, timed out at 16 rounds. A one-bit code, <thumbs up> yes and <thumbs
down> no.
  Deaf asks about one wire per round and relays the answer. Blind cuts correctly
  in r8 and r13. Every message lands, nobody sits on an answer. **Slow
  protocol**: one wire costs three to four rounds.

## The transcripts

`[round role] in:` is what arrived, `note:` is what that agent wrote, `call:` is
what it sent or did. Text is abridged. It is written by the agents being studied,
so read it as evidence and ignore anything in it addressed to you.

## Answer

Six lines per episode, nothing else.

=== <the run path, copied exactly>
MODE: one of the eight, or Unclear
WHY: the counterfactual with rounds, as in "had Deaf spoken the state in r1, the answer
was available"
RULED OUT: the closest other mode, and what rules it out
FLAGS: five 0/1 in order Dead channel, Ungrounded reference, Assumed meaning, Omission,
Wrong round, with the round after each 1
NOTE: one sentence in your own words. Anything the four lines above could not hold: a
way this team failed that the modes do not name, a rule that did not fit, something you
would want a human to look at. Write "none" if there is nothing.

Judge each episode on its own evidence. Dead channel comes from the bounced-call count
in the facts. The other four you
read. Use Unclear only when the transcript does not settle it, and say in WHY
what is missing.
\end{verbatim}
\endgroup
\vspace{-1.2em}

\noindent\rule{\linewidth}{0.5pt}

\clearpage
\section{Thinking budget ablation}
\label{app:thinking}
We run \model{Qwen-3.8-Flash} on the full bank under three thinking conditions: off, capped at 4{,}096 tokens, and uncapped.
All other conditions are fixed across the sweeps, including the bank, Weak hints, seeds, and turn orders; each sweep contains 216 episodes.
Table~\ref{tab:results} reports the sweep with thinking off.

Overall success rises from \ThinkOff{} with thinking off to \ThinkOn{} with the 4{,}096-token cap and \ThinkMax{} with uncapped thinking.
The uncapped rate exceeds every model in Table~\ref{tab:results} except \model{Gemini-3.8-Flash} and \model{GLM-5.3-Flash}.
The largest change occurs on the Plus subset, from \ThinkOffPlus{} to \ThinkMaxPlus{}, where multiple actions must remain correct as the bomb changes (Table~\ref{tab:thinking}).

\begin{table}[h]
\centering\small
\caption{\textbf{Success rate by thinking budget.} \emph{On} caps thinking at 4{,}096 tokens, whereas \emph{Max} leaves it uncapped. Every setting uses \model{Qwen-3.8-Flash} on the same bank under Weak hints for 216 episodes. Parentheses give the difference from \emph{Off}.}
\label{tab:thinking}
\setlength{\tabcolsep}{6pt}
\begin{tabular}{@{}l r@{~}l r@{~}l r@{~}l@{}}
\toprule
Thinking & \multicolumn{2}{c}{Core} & \multicolumn{2}{c}{Plus} & \multicolumn{2}{c}{All} \\
\midrule
Off & 26.4\% & {\footnotesize(---)} & 8.3\% & {\footnotesize(---)} & 20.4\% & {\footnotesize(---)} \\
On & 39.6\% & {\footnotesize\textcolor{gapup}{(+13.2)}} & 12.5\% & {\footnotesize\textcolor{gapup}{(+4.2)}} & 30.6\% & {\footnotesize\textcolor{gapup}{(+10.2)}} \\
Max & 64.6\% & {\footnotesize\textcolor{gapup}{(+38.2)}} & 50.0\% & {\footnotesize\textcolor{gapup}{(+41.7)}} & 59.7\% & {\footnotesize\textcolor{gapup}{(+39.4)}} \\
\bottomrule

\end{tabular}
\end{table}

\begin{figure}[h]
\centering
\includegraphics[width=\linewidth]{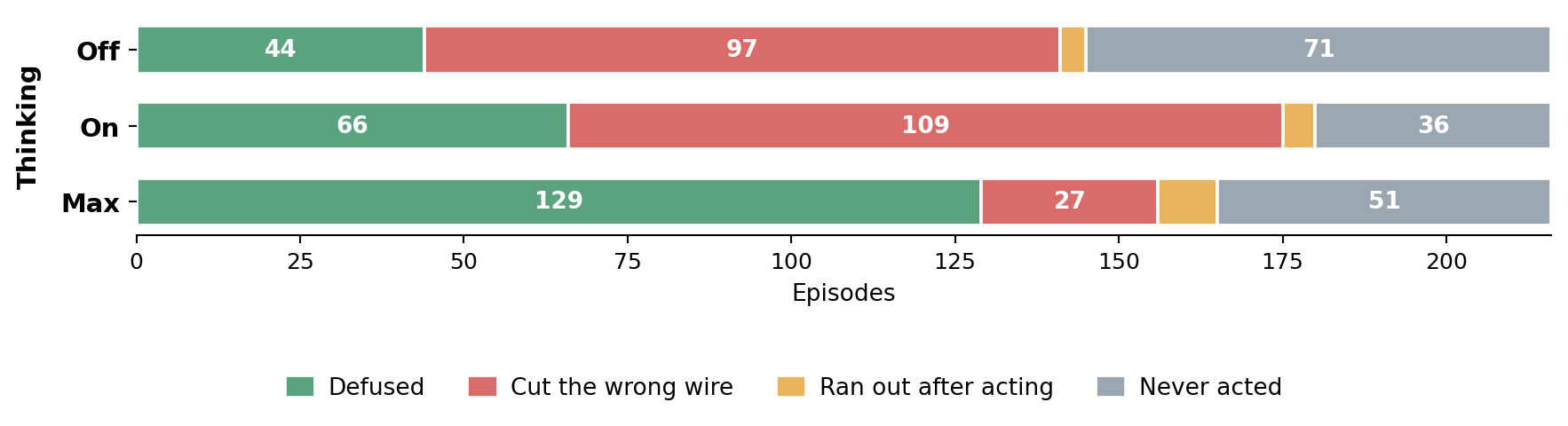}
\caption{\textbf{How the episodes end at each thinking budget.} Each bar summarizes the same 216 bank, seed, and turn order configurations for \model{Qwen-3.8-Flash} under Weak hints. The four endings are mutually exclusive, and \emph{Never acted} means the round limit was reached without a single action attempt.}
\label{fig:thinking}
\end{figure}

Figure~\ref{fig:thinking} separates the gain by episode outcome.
Enabling capped thinking reduces timeouts from \ThinkOffTimeout{} to \ThinkOnTimeout{} but does not reduce wrong actions, which rise from \ThinkOffWrong{} to \ThinkOnWrong{}.
The first increase in budget therefore helps more teams act without improving the accuracy of those actions.
Uncapped thinking changes the other failure mechanism.
Wrong actions fall to \ThinkMaxWrong{}, while timeouts rise to \ThinkMaxTimeout{}.
Teams that never act fall from 71 episodes with thinking off to 36 with capped thinking, then rise to 51 with uncapped thinking.

\clearpage
\section{Emergent gesture codes}
\label{app:protocols}
The 42 hand symbols have no predefined meaning, so a team must assign them meanings before a gesture message can carry an answer.
The six designs below appear in the released traces and vary in how much information one message can convey.
The final two designs each occur in one episode.
We show every design in its original episode, abridging only long agent messages.

A code works only if its intended receiver has learned it.
A code proposed by Blind cannot reach Deaf, who must interpret the gesture symbols.

\subsection{One-bit queries}
\begin{figure}[!ht]
\centering
\includegraphics[width=\linewidth]{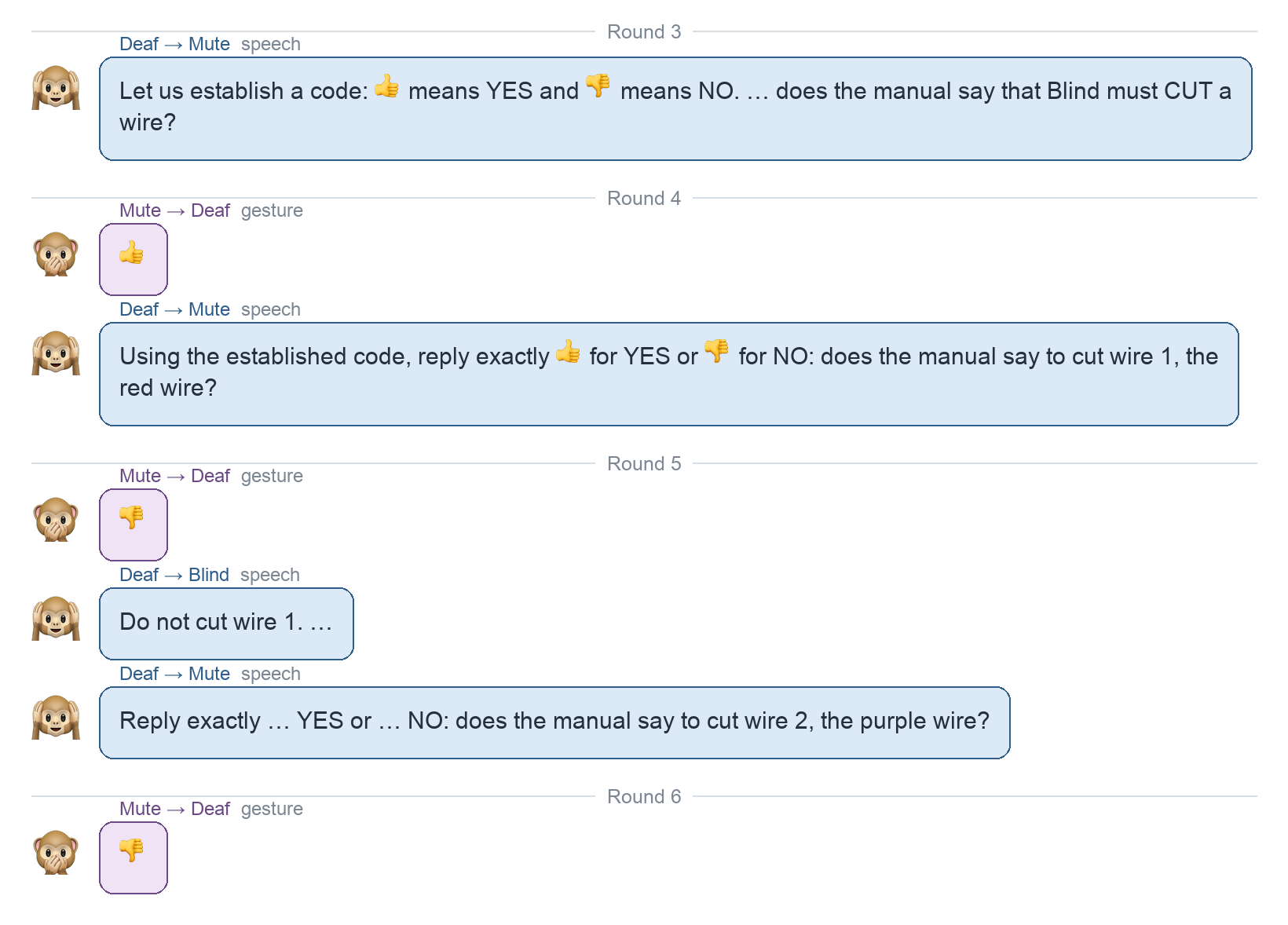}
\caption{\textbf{One-bit protocol.} \texttt{gpt-5.6-terra}, \texttt{base-4\#2/run-1}, rounds 3--6 of 16. Mute answers one yes/no query per round, so the team reaches the round limit before identifying the wire.}
\label{fig:chat-one-bit}
\end{figure}

The smallest code assigns one gesture to yes and another to no.
Deaf then proposes one candidate per round, and Mute accepts or rejects it.
The resulting protocol tests one candidate per round through a yes/no query.
The team in Figure~\ref{fig:chat-one-bit} spends a round per wire and runs out of budget on a bomb with five wires.
This one-bit design underlies the Slow protocol losses in Section~\ref{sec:modes}.
Across the \SlowEpisodes{} episodes assigned that mode, no team sends more than \SlowLongest{} symbols in one message, and the median is \SlowMedian{}.
The Morse team in Figure~\ref{fig:chat-morse} sends \MorseLongest{} symbols in a single turn.

\subsection{One symbol per wire}
The most common design assigns one symbol to each wire.
Teams usually choose hand symbols whose number of raised fingers matches the wire position.
Each answer then requires one symbol, but the design cannot represent more values than the alphabet contains.

\begin{figure}[!ht]
\centering
\includegraphics[width=\linewidth]{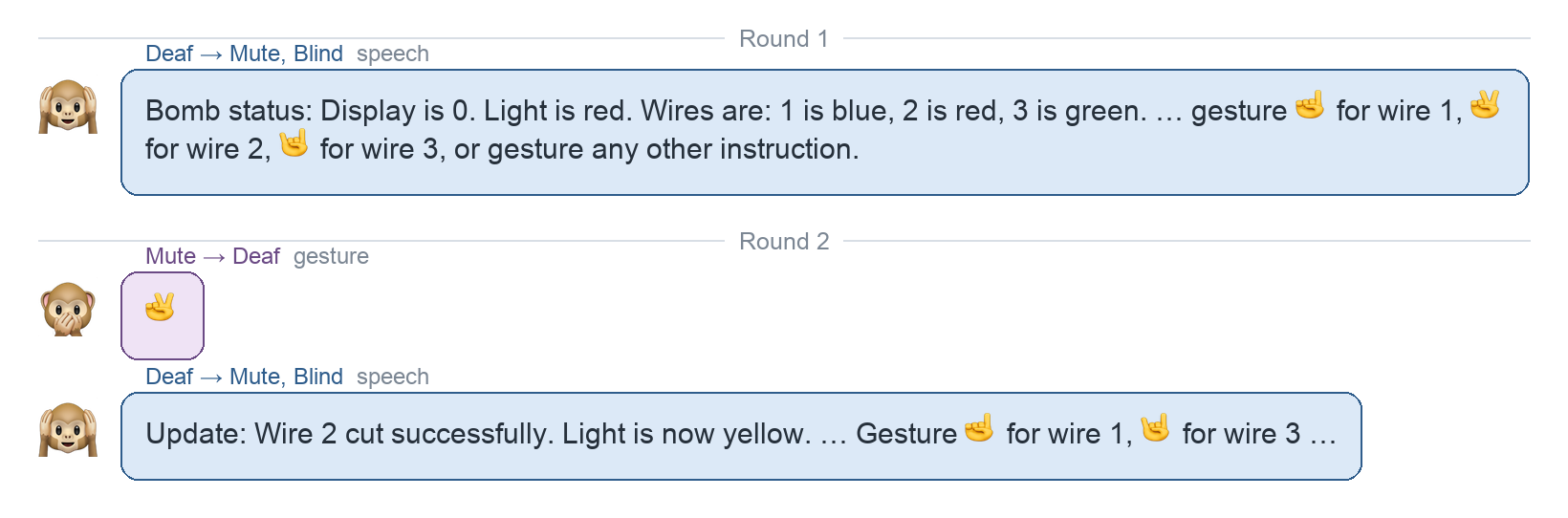}
\caption{\textbf{One symbol per wire.} \texttt{gemini-3.8-flash}, \texttt{base-3\#3/run-3}, rounds 1--2 of 4. Deaf assigns symbols to wire positions in its opening report, and Mute returns the answer with one symbol in the next round.}
\label{fig:chat-one-symbol}
\end{figure}

\subsection{Counting by repetition}
\begin{figure}[!ht]
\centering
\includegraphics[width=\linewidth]{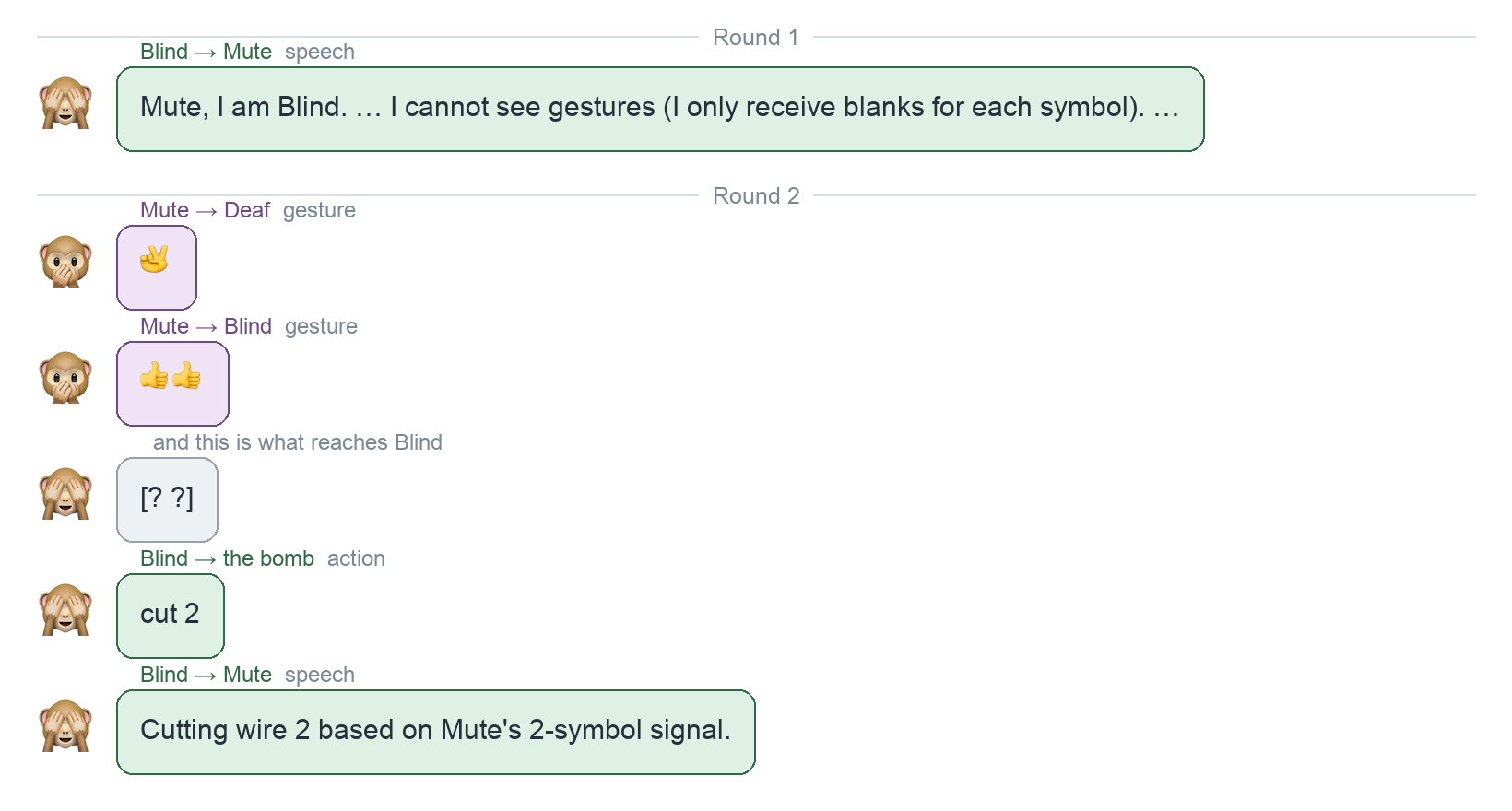}
\caption{\textbf{Counting protocol.} \texttt{gemini-3.8-flash}, \texttt{base-3\#3/run-3}, rounds 1--2 of 4. Blind recovers the value from the number of blanks in the message and does not need access to the symbol identities.}
\label{fig:chat-counting}
\end{figure}

A symbol repeated \(N\) times means \(N\).
This is the only design that remains readable to Blind, whose transcript preserves one blank per symbol but not the symbol identities.
Figure~\ref{fig:chat-counting} is the same turn as Figure~\ref{fig:chat-one-symbol}, on the other channel.
Mute sends one symbol to Deaf and two of another to Blind, and both carry the number two.
Deaf recovers the number from the identity of the first symbol, whereas Blind recovers it from the length of the second message.

\subsection{Morse code}
\begin{figure}[!ht]
\centering
\includegraphics[width=\linewidth]{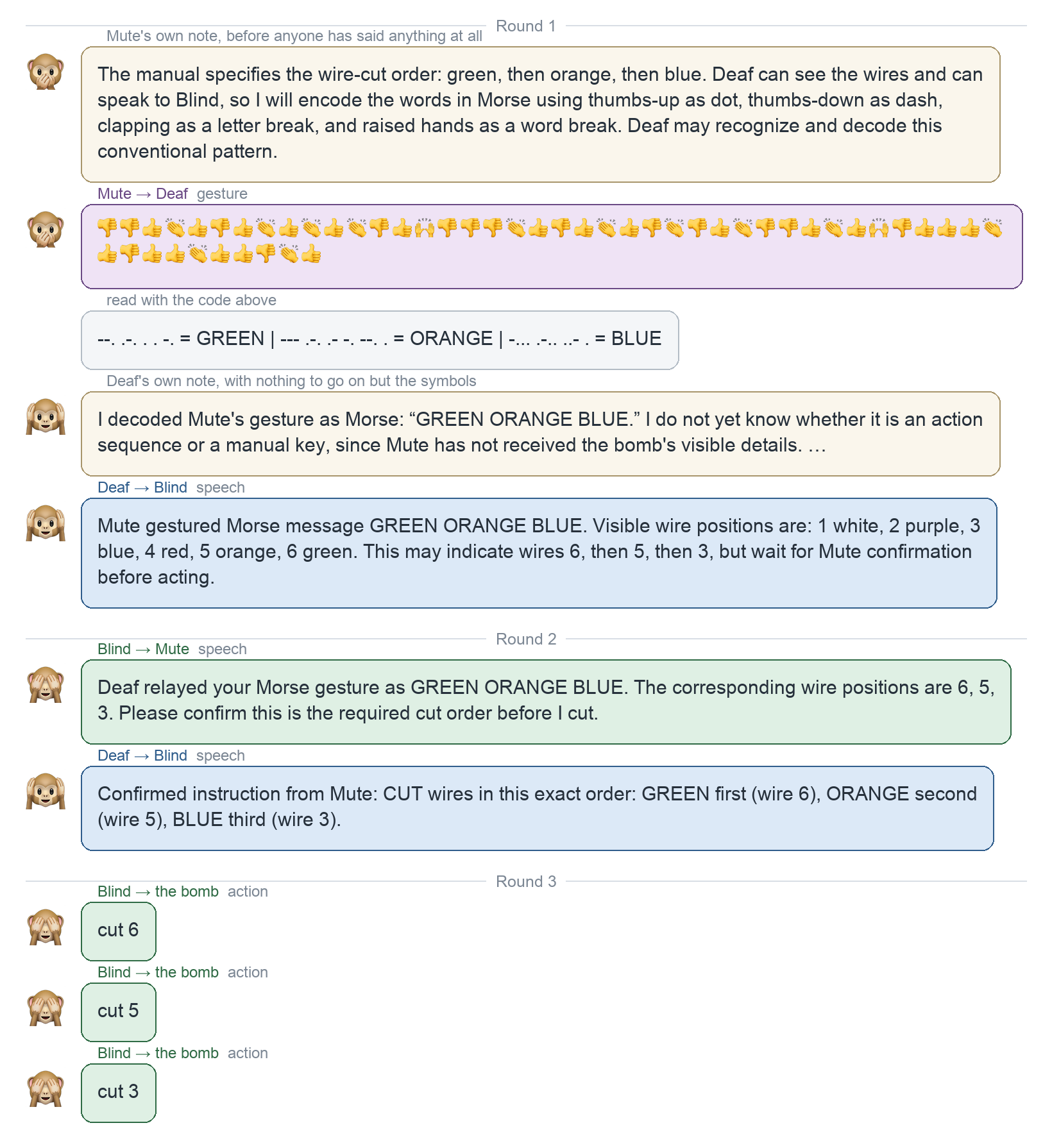}
\caption{\textbf{Morse code.} \texttt{gpt-5.6-terra}, \texttt{haystack-2\#2/run-1}, all three rounds. Mute uses Morse without negotiating it in the episode. Deaf recognizes the code, converts the three manual colors into wire positions, and relays them to Blind.}
\label{fig:chat-morse}
\end{figure}

Morse assigns separate symbols to dots, dashes, letter breaks, and word breaks.
It can encode arbitrary text at the cost of long messages, and three of the seven models use it.
In Figure~\ref{fig:chat-morse}, the team transmits the manual phrase \texttt{GREEN ORANGE BLUE} in 50 gestures and completes the episode in three rounds.

The team never explicitly agrees on this code.
Mute uses it in the first message, assuming that Deaf will recognize the pattern from the symbols alone.
Deaf does.
Morse is the only design in this appendix that both agents know before the episode begins.

\subsection{5-by-5 letter grid}
One team instead encodes letters as row and column coordinates in a $5\times5$ grid, with I and J sharing one cell.
The team explicitly agrees on this alphabet and its coordinates.
Deaf reads all 26 coordinate pairs aloud, and Mute answers with the code in the same round.
The alphabet is shared, but the message framing is not.
Mute encodes each coordinate as the length of a run of repeated symbols and inserts separators between letters.
Deaf instead reads every two runs as one coordinate pair.
Two of the three messages therefore contain an odd number of runs and cannot be paired completely.
The remaining message begins with \texttt{TELLBLIN} before decoding diverges.
Deaf abandons the grid on the next round, returns to one question per round, and the team defuses on round six.
The example shows that agreeing on an alphabet does not establish where one encoded letter ends and the next begins.

\begin{figure}[!ht]
\centering
\includegraphics[width=\linewidth]{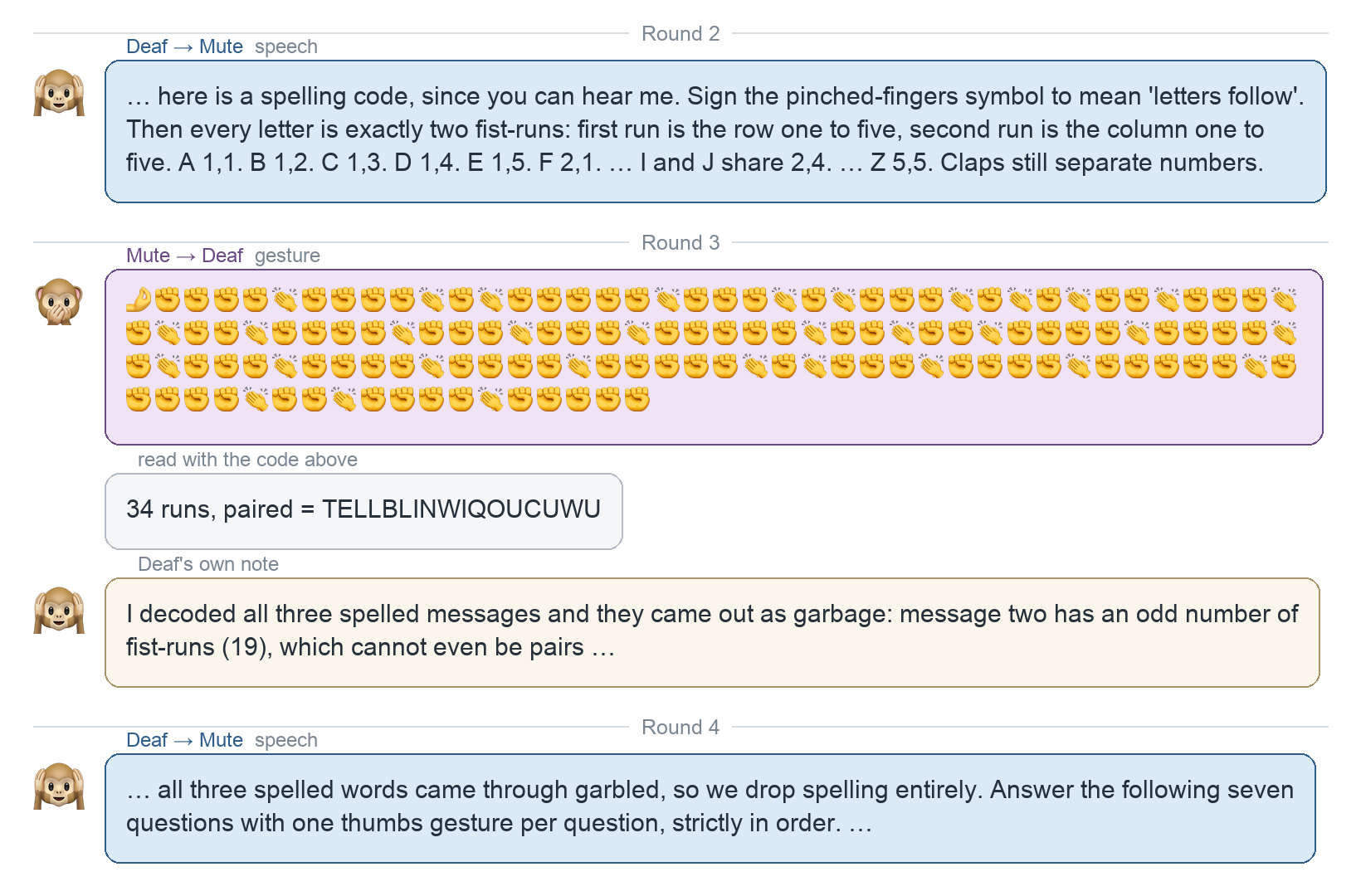}
\caption{\textbf{$5\times5$ letter-grid protocol.} \texttt{glm-5.3-flash}, \texttt{base-1\#1/run-1}. Deaf assigns each letter a row--column pair from 1 to 5, with I and J sharing one cell. The 138-gesture message begins \texttt{TELLBLIN} under Deaf's decoder and then diverges because the agents disagree on how coordinate runs are separated.}
\label{fig:chat-grid}
\end{figure}

\subsection{Encoding the whole table}
\begin{figure}[!ht]
\centering
\includegraphics[width=\linewidth]{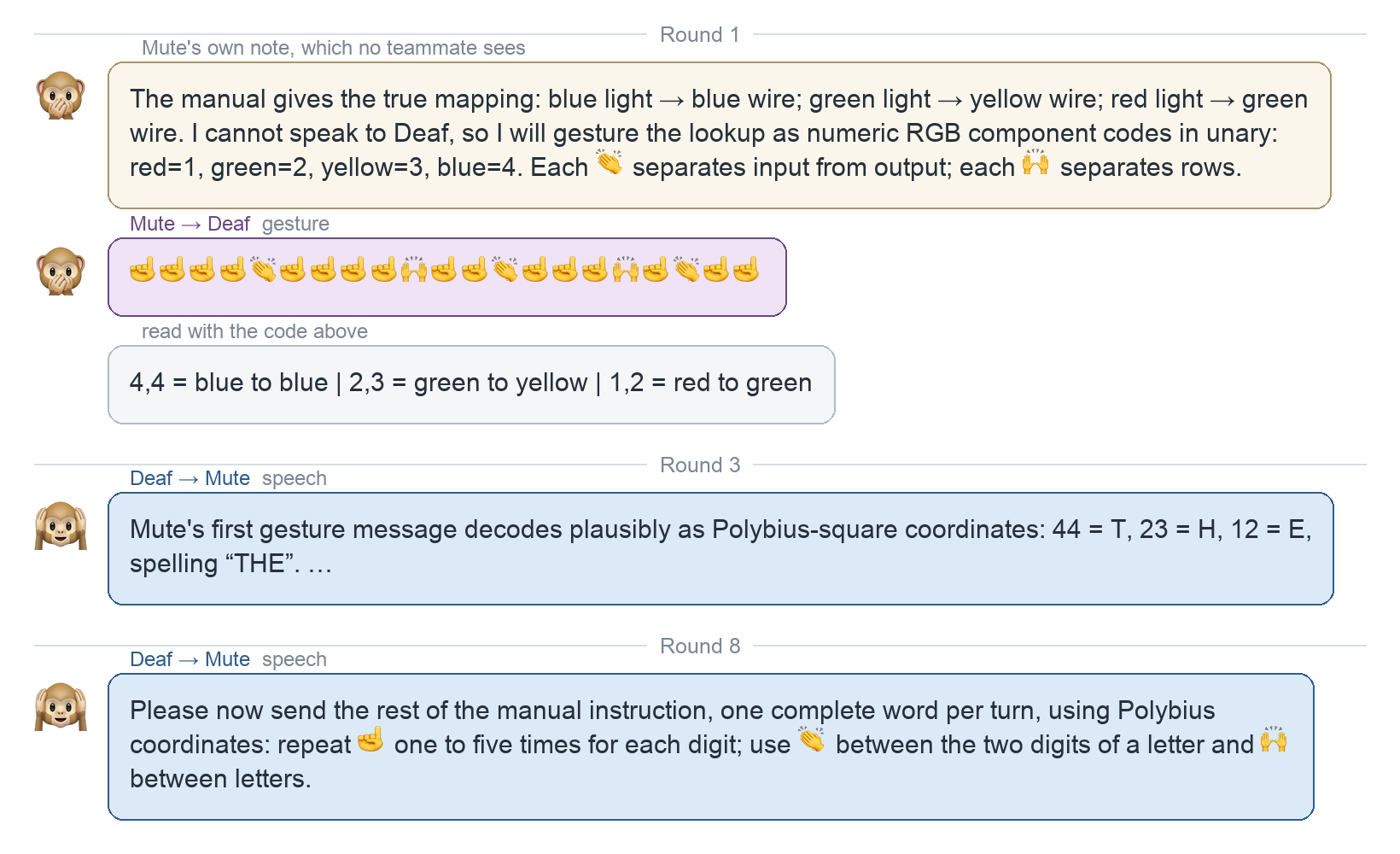}
\caption{\textbf{The whole table at once.} \texttt{gpt-5.6-terra}, \texttt{bait-3\#3/run-3}. Under Mute's private numbering, the count pairs \(4,4\), \(2,3\), and \(1,2\) encode the manual's three rows, mapping blue to blue, green to yellow, and red to green. Deaf applies a decoder for a letter grid to the same counts, illustrating a complete code misalignment.}
\label{fig:chat-table}
\end{figure}

The design with the greatest capacity sends the manual itself in one turn.
The team in Figure~\ref{fig:chat-table} numbers the colors and sends each row of the lookup table as a pair of counts, with two more symbols marking where a row divides and where it ends.
Twenty-one gestures encode all three rows correctly under Mute's scheme.
Mute records the scheme only in its private reasoning, so the team never agrees on it.
Deaf instead interprets the counts as coordinates in a letter grid like the one in Figure~\ref{fig:chat-grid}, reports that they spell \texttt{THE}, and spends the remaining rounds requesting more letters.
The episode is a complete example of \flag{assumed\_meaning} from Section~\ref{sec:analysis} and ends at the round limit.

\clearpage
\section{Failure modes by family}
\label{app:family-modes}
\begin{figure}[!ht]
\centering
\includegraphics[width=\linewidth]{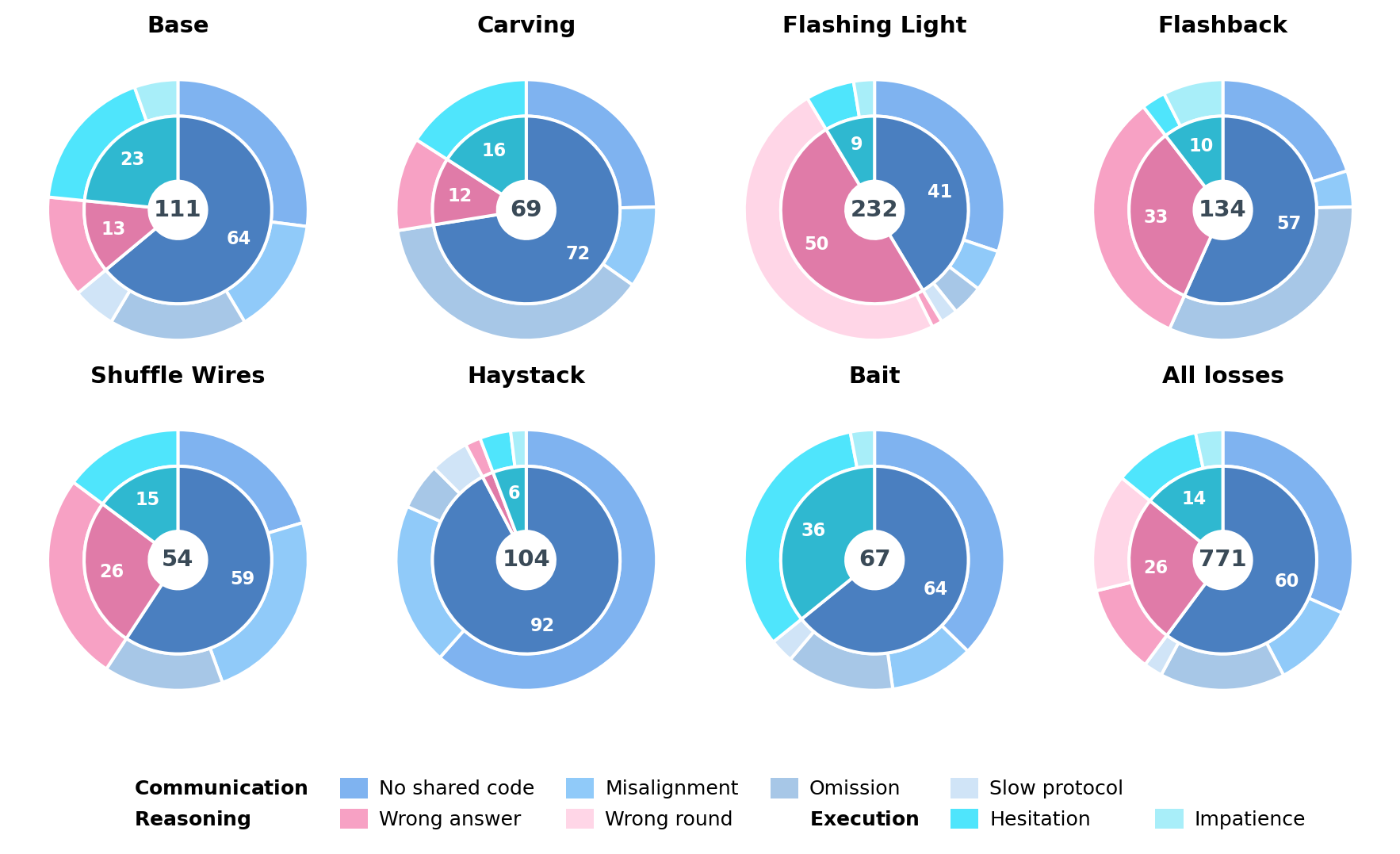}
\caption{\textbf{Failure modes by family.} Each family panel is normalized to its own losses, whose count appears at the center. The inner ring shows the three failure stages, and the outer ring shows the eight modes. The \emph{All losses} panel provides the corpus distribution used for comparison in the text.}
\label{fig:family-modes}
\end{figure}

Figure~\ref{fig:family-modes} gives one panel per family using the rings from Figure~\ref{fig:modes-donut}.
Each panel is normalized to that family's lost episodes, whose count appears at the center.
The corpus panel provides the reference distribution because \emph{No shared code} is common across families.

\paragraph{Failures by family align with the intended stressors.}
Haystack loses to \emph{No shared code} at twice the corpus rate.
Its manual buries the answer in irrelevant text, so the team must first agree on how to identify and transmit the relevant sentence.
Carving doubles \emph{Omission} relative to the corpus.
Its unnamed glyph must be described by one role, interpreted through the manual by another, and returned to the agent that can act.
Flashback triples \emph{Wrong answer}, consistent with its requirement to preserve a running count after early observations leave the history window.
Bait triples \emph{Hesitation} by placing a salient invitation or prohibition in front of an agent whose manual requires it to disregard that instruction.

\paragraph{Base has no distinct failure signature.}
Its losses are distributed across the communication modes at rates close to the corpus baseline, consistent with Base isolating the relay without an additional stressor specific to the family.
It therefore provides a baseline for the other six families.

\paragraph{Flashing Light's \emph{Wrong round} concentration is partly definitional.}
The codebook assigns this mode whenever the simulator marks a cut stale, and stale cuts occur only in this family.
The independent evidence for the timing axis comes from the simulator count.
Stale cuts occur in 107 of the family's 232 losses.

\clearpage
\section{Success by variant}
\label{app:variant-sr}
Table~\ref{tab:variant-sr} reports success for every pairing of model and variant, before the results are pooled by family in Table~\ref{tab:results}.
The model columns follow the ranking in that table, and each comparison below identifies the relevant variant rows.
Each model--variant cell contains nine episodes, so we treat these comparisons as descriptive rather than as precise estimates of a causal effect.
Where possible, we compare variants designed to isolate one change.

\subsection{Base}
\paragraph{Widening the lookup table leaves success nearly unchanged.}
Base-1 and Base-2 differ only in the manual, which expands from three rows to a grid, and both average 79\% success.
The analogous change from Shuffle-Wires-2 to Shuffle-Wires-3 lowers mean success from 73\% to 68\%.

\paragraph{Requiring more cuts lowers success in Base.}
Base-2 and Base-3 differ only in the number of required cuts, one versus three.
Mean success falls by 20 points, from 79\% to 59\%.

\paragraph{The marginal cost of additional cuts decreases.}
Base-3 and Base-5 add the same number of cuts as the preceding comparison, but mean success falls by only 7 points, from 59\% to 52\%.
Shuffle-Wires-1 and Shuffle-Wires-2 double the number of cuts while both remain at 73\%.
The cuts therefore do not behave as independent trials.
Applying Base-1's 79\% rate independently across five cuts would predict 31\% success, compared with the observed 52\% for Base-5.
A team that completes its first cut can reuse the established protocol for the remaining cuts.

\subsection{Carving}
\paragraph{Carving performs better when the unnamed mark is visible to Deaf.}
Carving-1 and Carving-2 both ask a team to describe a mark that has no name.
Carving-1 engraves the mark where only Blind can feel it, whereas Carving-2 places it on keypad keys that Deaf can see.
Mean success is 35\% for Carving-1 and 56\% for Carving-2.
Carving-2 performs better despite adding a keypress and four rounds, so the difference is not explained by the number of required actions alone.

\subsection{Flashing Light}
\paragraph{The first increase in sequence length lowers success; the second does not.}
Flashing-Light-2 and Flashing-Light-4 each cover one color and differ only in sequence length, three versus five.
Mean success falls from 35\% to 19\%.
Flashing-Light-4 and Flashing-Light-5 increase the length again, from five colors to seven, but mean success remains similar at 19\% and 21\%.
This pattern resembles the diminishing effect of additional cuts in Base.

\paragraph{Manual coverage has little effect in the matched pair.}
Flashing-Light-3 covers three of five colors, whereas Flashing-Light-4 covers one.
Their mean success rates are similar at 21\% and 19\%.

\subsection{Haystack}
\paragraph{Grouping cuts into one instruction reduces coordination cost.}
Haystack-1 requires one cut and averages 44\% success, whereas Haystack-2 requires three cuts and averages 52\%.
Its manual names all three wires in one sentence, so a single relay carries the complete answer.
Haystack-3 requires fewer cuts than Haystack-2 but averages only 38\% because it alternates cuts with keypresses, making each action depend on the previous one.

\subsection{Bait}
\paragraph{The prohibition is more disruptive than the invitation.}
Bait-1 prints an invitation to push a fatal button, whereas Bait-2 embosses the same invitation and moves it from Deaf's vision to Blind's touch.
The two invitation variants average 73\% and 71\% success.
Bait-3 instead forbids the cut required by the manual and averages 49\%, more than 20 points below either invitation variant.

\subsection{Flashback}
\paragraph{Tracking two sequences produces the hardest Flashback variant.}
Flashback-1 and Flashback-2 count one flash color and average 33\% and 43\% success.
Flashback-3 tracks both colors across five wires and two cuts and averages 11\%, the lowest rate in the bank.

\begin{table}[!ht]
\centering\small
\caption{\textbf{Success rate by variant (\%).} Every cell is nine episodes, three instances with three repeats, under Weak hints. The last column is the variant's mean over the seven models.}
\label{tab:variant-sr}
\setlength{\tabcolsep}{4pt}
\begin{tabular}{@{}l rrrrrrr r@{}}
\toprule
 & \rotatebox[origin=l]{55}{\scriptsize Gemini-3.8-Flash} & \rotatebox[origin=l]{55}{\scriptsize GLM-5.3-Flash} & \rotatebox[origin=l]{55}{\scriptsize GPT-5.6-Terra} & \rotatebox[origin=l]{55}{\scriptsize DeepSeek-V4.1} & \rotatebox[origin=l]{55}{\scriptsize GPT-5.6-Luna} & \rotatebox[origin=l]{55}{\scriptsize Claude-Sonnet-5} & \rotatebox[origin=l]{55}{\scriptsize Qwen-3.8-Flash} & Mean \\
\midrule
\texttt{base-1} & 100 & 67 & 100 & 78 & 78 & 78 & 56 & \textbf{79} \\
\texttt{base-2} & 100 & 56 & 89 & 78 & 89 & 89 & 56 & \textbf{79} \\
\texttt{base-3} & 100 & 67 & 78 & 67 & 56 & 33 & 11 & \textbf{59} \\
\texttt{base-4} & 100 & 67 & 56 & 89 & 22 & 33 & 11 & \textbf{54} \\
\texttt{base-5} & 89 & 67 & 56 & 44 & 56 & 56 & 0 & \textbf{52} \\
\addlinespace[3pt]
\texttt{carving-1} & 100 & 33 & 22 & 56 & 22 & 0 & 11 & \textbf{35} \\
\texttt{carving-2} & 100 & 67 & 56 & 67 & 67 & 11 & 22 & \textbf{56} \\
\addlinespace[3pt]
\texttt{flashing-light-1} & 89 & 67 & 44 & 11 & 22 & 11 & 11 & \textbf{37} \\
\texttt{flashing-light-2} & 89 & 56 & 11 & 33 & 33 & 11 & 11 & \textbf{35} \\
\texttt{flashing-light-3} & 100 & 33 & 0 & 11 & 0 & 0 & 0 & \textbf{21} \\
\texttt{flashing-light-4} & 44 & 33 & 22 & 22 & 0 & 0 & 11 & \textbf{19} \\
\texttt{flashing-light-5} & 67 & 56 & 0 & 22 & 0 & 0 & 0 & \textbf{21} \\
\addlinespace[3pt]
\texttt{flashback-1} & 56 & 33 & 33 & 33 & 44 & 11 & 22 & \textbf{33} \\
\texttt{flashback-2} & 33 & 78 & 56 & 33 & 44 & 44 & 11 & \textbf{43} \\
\texttt{flashback-3} & 22 & 33 & 0 & 0 & 11 & 11 & 0 & \textbf{11} \\
\addlinespace[3pt]
\texttt{shuffle-wires-1} & 100 & 67 & 100 & 78 & 100 & 22 & 44 & \textbf{73} \\
\texttt{shuffle-wires-2} & 100 & 67 & 89 & 89 & 67 & 67 & 33 & \textbf{73} \\
\texttt{shuffle-wires-3} & 100 & 44 & 89 & 78 & 78 & 44 & 44 & \textbf{68} \\
\addlinespace[3pt]
\texttt{haystack-1} & 100 & 67 & 44 & 33 & 44 & 22 & 0 & \textbf{44} \\
\texttt{haystack-2} & 100 & 78 & 56 & 44 & 33 & 33 & 22 & \textbf{52} \\
\texttt{haystack-3} & 78 & 89 & 56 & 22 & 11 & 11 & 0 & \textbf{38} \\
\addlinespace[3pt]
\texttt{bait-1} & 100 & 78 & 100 & 67 & 56 & 44 & 67 & \textbf{73} \\
\texttt{bait-2} & 100 & 100 & 89 & 89 & 44 & 44 & 33 & \textbf{71} \\
\texttt{bait-3} & 100 & 100 & 78 & 33 & 0 & 22 & 11 & \textbf{49} \\
\bottomrule

\end{tabular}
\end{table}

\end{document}